\documentclass{article} 
\usepackage{fourier}
\usepackage{amsmath,amssymb,amsthm}
\usepackage{fullpage}
\usepackage{url}
\usepackage[numbers]{natbib}
\usepackage{color}

\usepackage{algorithm,algorithmic}
\usepackage[colorlinks=true,linkcolor=blue]{hyperref}

\newtheorem{theorem}{Theorem}
\newtheorem{lemma}[theorem]{Lemma}

\newtheorem{corollary}[theorem]{Corollary}

\newtheorem{proposition}[theorem]{Proposition}
\newtheorem{example}{Example}
\newtheorem{remark}{Remark}

\theoremstyle{definition}
\newtheorem{definition}{Definition}

\DeclareMathOperator{\Ex}{\vphantom{p}\mathbb{E}}
\let\inf\undef
\DeclareMathOperator*{\inf}{\vphantom{p}inf}
\let\sup\undef
\DeclareMathOperator*{\sup}{\vphantom{p}sup}

\newcommand{\mbf}[1]{\mathbf{#1}}

\newcommand{\mrm}[1]{\mathrm{#1}}

\newcommand{\norm}[1]{\left\|#1\right\|}

\newcommand{\sign}{\mrm{sign}}
\newcommand{\argmin}[1]{\underset{#1}{\mrm{argmin}} \ }

\newcommand{\reals}{\mathbb{R}}
\newcommand{\E}[1]{\mathbb{E}\left[ #1 \right]} 
\newcommand{\En}{\mathbb{E}}  
\newcommand{\Es}[2]{\Ex_{#1}\left[ #2 \right]} 

\newcommand{\inner}[1]{\left\langle #1 \right\rangle}
\newcommand{\ip}[2]{\left<#1,#2\right>}

\newcommand\s{\mathbf{s}}
\newcommand\w{\mathbf{w}}
\newcommand\x{\mathbf{x}}

\newcommand\z{\mathbf{z}}

\renewcommand\v{\mathbf{v}}

\newcommand\cN{\mathcal{N}}

\newcommand\X{\mathcal{X}}

\newcommand\Y{\mathcal{Y}}
\newcommand\Z{\mathcal{Z}}
\newcommand\F{\mathcal{F}}
\newcommand\G{\mathcal{G}}
\newcommand\cH{\mathcal{H}}

\newcommand\fat{\mathrm{fat}}

\newcommand\Rad{\mathfrak{R}}

\newcommand\Reg{\mbf{Reg}}

\newcommand{\Rel}[2]{\mbf{Rel}_{#1}\left(#2 \right)}

\newcommand{\bmu}{\ensuremath{\boldsymbol{\mu}}}
\newcommand{\bta}{\ensuremath{\boldsymbol{\eta}}}
\include{multiminimax}

\title{Online Nonparametric Regression}
\author{Alexander Rakhlin \\ University of Pennsylvania \and Karthik Sridharan \\ University of Pennsylvania}

\begin{document}

\maketitle

\begin{abstract}
	We establish optimal rates for online regression for arbitrary classes of regression functions in terms of the sequential entropy introduced in \citep{RakSriTew10}. The optimal rates are shown to exhibit a phase transition analogous to the i.i.d./statistical learning case, studied in \citep{RakSriTsy14}. In the frequently encountered situation when sequential entropy and i.i.d. empirical entropy match, our results point to the interesting phenomenon that the rates for statistical learning with squared loss and online nonparametric regression are the same.

	In addition to a non-algorithmic study of minimax regret, we exhibit a generic forecaster that enjoys the established optimal rates. We also provide a recipe for designing online regression algorithms that can be computationally efficient. We illustrate the techniques by deriving existing and new forecasters for the case of finite experts and for online linear regression.
\end{abstract}

\section{Introduction}

Within the online regression framework, data $(x_1,y_1), \ldots, (x_n,y_n),\ldots$ arrive in a stream, and we are tasked with sequentially predicting each next response $y_t$ given the current $x_t$ and the data $\{(x_i,y_i)\}_{i=1}^{t-1}$ observed thus far. Let $\hat{y}_t$ denote our prediction, and let the quality of this forecast be evaluated via square loss $(\hat{y}_t-y_t)^2$. Within the field of time series analysis, it is assumed that data are generated according to some model. The parameters of the model can then be estimated from data, leveraging the laws of probability. Alternatively, in the \emph{competitive approach}, studied within the field of online learning, the aim is to develop a prediction method that does not assume a generative process of the data \citep{PLG}. The problem is then formulated as that of minimizing regret 
\begin{align}
	\label{eq:standard_regret}
	\sum_{t=1}^n (\hat{y}_t-y_t)^2 - \inf_{f\in\F} \sum_{t=1}^n (f(x_t)-y_t)^2
\end{align}
with respect to some benchmark class of functions $\F$. This class encodes our prior belief about the family of regression functions that we expect to perform well on the sequence. Notably, an upper bound on regret is required to hold for all sequences. 

In the past twenty years, progress in online regression for arbitrary sequences, starting with the paper of Foster \cite{f-pwc-91}, has been almost exclusively on finite-dimensional \emph{linear} regression (an incomplete list includes \citep{Vovk98, KivWar97, v-cos-01, cb-atgbaolr-99, acbg-ascola-02, AzouryWarmuth01, gerchinovitz2013sparsity}). This is to be contrasted with Statistics, where regression has been studied for rich (nonparametric) classes of functions. Important exceptions to this limitation in the online regression framework -- and works that partly motivated the present findings -- are the papers of Vovk \citep{Vovk07wild,Vovk06metric,vovk2006hilbert}. Vovk considers regression with large classes, such as subsets of a Besov or Sobolev space, and remarks that there appears to be two distinct approaches to obtaining the upper bounds in online competitive regression. The first approach, which Vovk terms Defensive Forecasting, exploits uniform convexity of the space, while the second -- an aggregating technique (such as the Exponential Weights Algorithm) -- is based on the metric entropy of the space. Interestingly, the two seemingly different approaches yield distinct upper bounds, based on the respective properties of the space. In particular, Vovk asks whether there is a unified view of these techniques. The present paper addresses these questions and establishes optimal performance for online regression. 

Since most work in online learning is algorithmic, the boundaries of what can be proved are defined by the regret minimization algorithms one can find. One of the main algorithmic workhorses is the aggregating procedure mentioned above. However, the difficulty in using an aggregating procedure beyond simple parametric classes (e.g. subsets of $\reals^d$) lies in the need for a ``pointwise'' cover of the set of functions -- that is, a cover in the supremum norm on the underlying space of covariates (see Remark~\ref{rem:vovk}). The same difficulty arises when one uses PAC-Bayesian bounds \citep{audibert2009fast} that, at the end of the day, require a volumetric argument. Notably, this difficulty has been overcome in statistical learning, where it has long been recognized (since the work of Vapnik and Chervonenkis) that it is sufficient to consider an \emph{empirical} cover of the class -- a potentially much smaller quantity. Such an empirical entropy is necessarily finite, and its growth with $n$ is one of the key complexity measures for i.i.d. learning. In particular, the recent work of \cite{RakSriTsy14} shows that the behavior of empirical entropy characterizes the optimal rates for i.i.d. learning with square loss. To mimic this development, it appears that we need to understand empirical covering numbers in the sequential prediction framework.

Sequential analogues of covering numbers, combinatorial parameters, and the Rademacher complexity have been recently introduced in \citep{RakSriTew14}. These complexity measures were shown to both upper and lower bound minimax regret of online learning with absolute loss for arbitrary classes of functions. These rates, however, are not correct for the square loss case. Consider, for instance, finite-dimensional regression, where the behavior of minimax regret is known to be logarithmic in $n$; the Rademacher rate, however, cannot yield rates faster than $\sqrt{n}$. A hint as to how to modify the analysis for ``curved'' losses appears in the paper of \cite{cesa1999minimax} where the authors derived rates for log-loss via a two-level procedure: the set of densities is first partitioned into small balls of a critical radius $\gamma$; a minimax algorithm is employed on each of these small balls; and an overarching aggregating procedure combines these algorithms. Regret within each small ball is upper bounded by classical Dudley entropy integral (with respect to a pointwise metric) defined up to the $\gamma$ radius. The main technical difficulty in this paper is to prove a similar statement using ``empirical'' sequential covering numbers.\footnote{While we develop our results for square loss, similar statements hold for much more general losses, as will be shown in the full version of this paper.} 

Interestingly, our results imply the same phase transition as the one exhibited in \citep{RakSriTew14} for i.i.d. learning with square loss. More precisely, under the assumption of the $O(\beta^{-p})$ behavior of sequential entropy, the minimax regret normalized by time horizon $n$ decays as $n^{-\frac{2}{2+p}}$ if $p\in(0,2]$, and as $n^{-1/p}$ for $p\geq 2$. We prove lower bounds that match up to a logarithmic factor, establishing that the phase transition is real. Even more surprisingly, it follows that, under a mild assumption that sequential Rademacher complexity of $\F$ behaves similarly to its i.i.d. cousin, \emph{the rates of minimax regret in online regression with arbitrary sequences match, up to a logarithmic factor, those in the i.i.d. setting of Statistical Learning}. This phenomenon has been noticed for some parametric classes by various authors (e.g. \citep{cbfhhsw-huea-97}). The phenomenon is even more striking given the simple fact that one may convert the regret statement, that holds for all sequences, into an i.i.d. guarantee. Thus, in particular, we recover the result of \citep{RakSriTsy14} through completely different techniques. Since in many situations, one obtains optimal rates for i.i.d. learning from a regret statement, the relaxation framework of \citep{rakhlin2012relax} provides a toolkit for developing improper learning algorithms in the i.i.d. scenario. 

After characterizing minimax rates for online regression, we turn to the question of developing algorithms. We first show that an algorithm based on the Rademacher relaxation is admissible (see \citep{rakhlin2012relax}) and yields the rates derived in a non-constructive manner in the first part of the paper. This algorithm is not generally computationally feasible, but, in particular, does achieve optimal rates, improving on those exhibited by Vovk \cite{Vovk06metric} for Besov spaces. We show that further relaxations in finite dimensional space lead to the famous Vovk-Azoury-Warmuth forecaster. For illustration purposes, we also derive a prediction method for finite class $\F$.

\section{Background}

Let $\X$ be some set of covariates, and let $\F$ be a class of functions $\X\to[-1,1]=\Y$. We study the online regression scenario where on round $t\in \{1,\ldots,n\}$, $x_t\in\X$ is revealed to the learner who subsequently makes a prediction $\hat{y}_t\in\reals$; Nature then reveals\footnote{The assumption of bounded responses can be removed by standard truncation arguments (see e.g. \citep{gerchinovitz2013adaptive}). } $y_t\in [-B,B]$. Instead of \eqref{eq:standard_regret}, we consider a slightly modified notion of regret
\begin{align}
	\label{eq:alpha-regret}
	(1-\alpha)\sum_{t=1}^n (\hat{y}_t-y_t)^2 - \inf_{f\in\F}\sum_{t=1}^n (f(x_t)-y_t)^2
\end{align}
for some $\alpha\in[0,1)$. It is well-known that an upper bound on such a regret notion leads to the so-called \emph{optimistic rates} which scale favorably with the cumulative loss 
$L^* = \inf_{f\in\F}\sum_{t=1}^n (f(x_t)-y_t)^2$ ~~\citep{acbg-ascola-02,srebro2010smoothness}. More precisely, suppose we show an upper bound of $U_1/\alpha + U_2$ on regret in \eqref{eq:alpha-regret}. Then regret in \eqref{eq:standard_regret} is upper bounded by
\begin{align}
	\label{eq:generic_Lstar}
	4\sqrt{L^* U_1}+12U_1+4U_2 
\end{align}
by considering the case $L^*\geq 4U_1$ and its converse.

Unlike most previous approaches to the study of online regression, we do not start from an algorithm, but instead directly work with minimax regret. We will be able to extract a (not necessarily efficient) algorithm after getting a handle on the minimax value. Let us introduce the notation that makes the minimax regret definition more concise. We use $\multiminimax{\cdots}_{t=1}^n$ to denote an  interleaved application of the operators inside repeated over  $t = 1\ldots n$ rounds. With this notation, the minimax regret of the online regression problem described earlier can be written as
\begin{align}
	\label{eq:def_value}
	V_n^\alpha = \multiminimax{\sup_{x_t}\inf_{\hat{y}_t}\sup_{y_t}}_{t=1}^n\left\{(1-\alpha)\sum_{t=1}^n (\hat{y}_t-y_t)^2 - \inf_{f\in\F}\sum_{t=1}^n (f(x_t)-y_t)^2 \right\}
\end{align}
where each $x_t$ ranges over $\X$ and $\hat{y}_t, y_t$ range over $[-B,B]$. The usual minimax regret notion is simply given when $\alpha =0$ as $V_n^0$.

As mentioned above, in the i.i.d. scenario it is possible to employ a notion of a cover based on a sample, thanks to the symmetrization technique. In the online prediction scenario, symmetrization is more subtle, and involves the notion of a binary tree, the smallest entity that captures the sequential nature of the problem. To this end, let us state a few definitions. A $\Z$-valued tree $\z$ of depth $n$ is a complete rooted binary tree with nodes labeled by elements of $\Z$. Equivalently, we think of $\z$ as $n$ labeling functions, where $\z_1$ is a constant label for the root, $\z_2(-1),\z_2(+1)\in \Z$ are the labels for the left and right children of the root, and so forth. Hence, for $\epsilon=(\epsilon_1,\ldots,\epsilon_n)\in \{\pm1\}^n$, $\z_t(\epsilon)=\z_t(\epsilon_1,\ldots, \epsilon_{t-1})\in \Z$ is the label of the node on the $t$-th level of the tree obtained by following the path $\epsilon$. For a function $g:\Z\to\reals$, $g(\z)$ is an $\reals$-valued tree with labeling functions $g\circ \z_t$ for level $t$ (or, in plain words, evaluation of $g$ on $\z$).

Next, let us define sequential covering numbers -- one of the key complexity measures of $\F$.
\begin{definition}[\cite{RakSriTew14}]
		A set $V$ of $\reals$-valued trees of depth $n$ forms a $\beta$-cover (with respect to the $\ell_q$ norm) of a function class $\F\subseteq \reals^\X$ on a given $\X$-valued tree $\x$ of depth $n$ if 
		$$\forall f\in\F, \forall \epsilon\in\{\pm1\}^n, \exists \v\in V ~~~~\mbox{s.t.}~~~~ \frac{1}{n}\sum_{t=1}^n |f(\x_t(\epsilon))-\v_t(\epsilon)|^q \leq \beta^q.$$
		A $\beta$-cover in the $\ell_\infty$ sense requires that $|f(\x_t(\epsilon))-\v_t(\epsilon)| \leq \beta$ for all $t\in[n]$. The size of the smallest $\beta$-cover is denoted by $\cN_q(\beta, \F, \x)$, and $\cN_q(\beta, \F, n) = \sup_{\x}\log \cN_q(\beta, \F, \x)$.
\end{definition}

We will refer to $\sup_{\x}\log \cN_q(\beta, \F, \x)$ as \emph{sequential entropy} of $\F$. In particular, we will study the behavior of $V_n^\alpha (\F)$ when sequential entropy grows polynomially\footnote{It is straightforward to allow constants in this definition, and we leave these details out for the sake of simplicity.} as the scale $\beta$ decreases:
\begin{align}
	\label{eq:entropy_growth}
	\log \cN_2(\beta, \F, n) = \beta^{-p}, ~~~~p>0.
\end{align}
We also consider the parametric ``$p=0$'' case when sequential covering itself behaves as  
\begin{align}
	\label{eq:vc_growth}
	\cN_2(\beta,\F,n) = \beta^{-d}
\end{align}
(e.g. linear regression in a bounded set in $\reals^d$). We remark that the $\ell_\infty$ cover is necessarily $n$-dependent, so the form we assume there is
\begin{align}
	\label{eq:vc_growth_infty}
	\cN_\infty(\beta,\F,n) = (n/\beta)^{-d} \ .
\end{align}

\section{Main Results}
We now state the main results of this paper. They follow from the more general technical statements of Lemmas~\ref{lem:expansion}, \ref{lem:reverse_dudley}, \ref{lem:reverse_dudley_with_alpha} and \ref{lem:finite_lemma_with_etas}. We normalize $V_n^\alpha$ by $n$ in order to make the rates comparable to those in statistical learning. Further, throughout the paper $C,c$ refer to constants that may depend on $B,p$. Their values can be found in the proofs.
\begin{theorem}
	\label{thm:upper}
	For a class $\F$ with sequential entropy growth $\log \cN_2(\beta, \F, n) \leq \beta^{-p}$,
	\begin{itemize}
		\item For $p> 2$, the minimax regret\footnote{For $p=2$, $\frac{1}{n}V^0_n \leq C\log(n)n^{-1/2}$.} is bounded as ~~$\frac{1}{n}V^0_n \leq Cn^{-1/p}$ 
		\item For $p\in (0,2)$, the minimax regret is bounded as ~~$\frac{1}{n}V^0_n \leq Cn^{-2/(2+p)}$
		\item For the parametric case \eqref{eq:vc_growth}, ~~$\frac{1}{n}V^0_n \leq C dn^{-1}\log(n)$
		\item For finite set $\F$,  ~~$\frac{1}{n}V^0_n \leq C n^{-1}\log|\F|$
	\end{itemize}
\end{theorem}
\begin{theorem}
	\label{thm:lower}
	The upper bounds of Theorem~\ref{thm:upper} are tight\footnote{The $\tilde{\Omega}(\cdot)$ notation suppresses logarithmic factors}:
	\begin{itemize}
		\item For $p\geq 2$, for any class $\F$ of uniformly bounded functions with a lower bound of $\beta^{-p}$ on sequential entropy growth, $\frac{1}{n}V^0_n \geq \tilde{\Omega}(n^{-1/p})$
		\item For $p\in(0,2]$, for any class $\F$ of uniformly bounded functions, there exists a slightly modified class $\F'$ with the same sequential entropy growth such that $\frac{1}{n}V^0_n \geq \tilde{\Omega}(n^{-2/(2+p)})$
		\item There exists a class $\F$ with the covering number as in \eqref{eq:vc_growth},  such that $\frac{1}{n}V^0_n \geq \Omega(dn^{-1}\log(n))$
	\end{itemize}
\end{theorem}

For the following theorem, we assume that $L^*$ is known a priori. Adaptivity to $L^*$ can be obtained through a doubling-type argument \citep{ShalevShwartz07thesis}.
\begin{theorem}
	\label{thm:optimistic}
	Additionally, the following optimistic rates hold for regret \eqref{eq:standard_regret}:
	\begin{itemize}
		\item For $p> 2$, regret is upper bounded by $C\sqrt{L^* n^{1-1/(p-1)}\log(n)}+ Cn^{1-1/(p-1)}\log(n)$
		\item For $p\in(0,2)$, regret is upper bounded by $C\sqrt{L^*\log(n)}+C\log(n)$. The bound gains an extra $\log(n)$ factor for $p=2$
		\item For the parametric case \eqref{eq:vc_growth_infty}, regret is upper bounded by $C\sqrt{L^*d\log(n)}+Cd\log(n)$
	\end{itemize}
	where $L^* =\inf_{f\in\F}\sum_{t=1}^n (f(x_t)-y_t)^2$. 
\end{theorem}

\begin{remark}
	The optimistic rate for $p>2$ appears to be slower than the hypothesized $\sqrt{L^*n^{1-2/p}}+n^{1-2/p}$ rate, and we leave the question of obtaining this rate as future work.
\end{remark}

\begin{remark}
	If we assume that $y_t$'s are drawn from distributions with bounded mean and subgaussian tails, the same upper bounds can be shown with an extra $\log(n)$ factor.
\end{remark}

Next, we prove the three theorems stated above. The proofs are of the ``plug-and-play style'': the overarching idea is that the optimal rates can be derived simply by assuming an appropriate control of sequential entropy, be it a parametric or a nonparametric class. 

\begin{proof}[\textbf{Proof of Theorem~\ref{thm:upper}}]
	We appeal to Eq.~\eqref{eq:val_upper_0_alpha} in Lemma~\ref{lem:expansion} below. Fix $\x,\bmu$ and let $\z$ denote the $\X\times\reals$-valued tree $(\x,\bmu)$. Define the class $\G=\{g_f: g_f(\z)= f(\x)-\bmu, f\in\F\}$. Observe that the values of $g_f$ outside of range of $\z$ are immaterial. Also note that the covering number of $\G$ on $\z$ coincides with the covering number of $\F$ on $\x$. Now, Lemma~\ref{lem:reverse_dudley} applied to this class $\G$, together with $\bta\equiv B$, yields
	\begin{align}
		\label{eq:bound1_no_alpha}
		V_n^0 \leq 32B^2 \log \cN_2(\gamma, \F, n)+ B\inf_{\rho\in(0,\gamma)} \left\{ 4\rho n + 12\sqrt{n}\int_{\rho}^\gamma \sqrt{\log\cN_2(\delta,\F,n)}d\delta \right\} 
	\end{align}
	We now evaluate the above upper bound for the $\beta^{-p}$ growth of sequential entropy at scale $\beta$. In particular, for the case $p>2$, we may choose $\gamma=1$ (maximum of the function) and $\rho=n^{-1/p}$. Then $\cN_2(B, \F, n)=1$ and the first term disappears. We are left with
	\begin{align*}
		B^{-1} V_n^0 
		\leq 4n^{1-p}+12\sqrt{n}\left[ \left(\frac{2}{2-p}\right)\delta^{(2-p)/2}\right]_{n^{-1/p}}^B 
		\leq 4n^{1-\frac{1}{p}}+\frac{24}{p-2} n^{-\frac{2-p}{2p}+\frac{1}{2}} = \left(4+\frac{24}{p-2}\right) n^{1-1/p}
	\end{align*}
	For the case $p\in(0,2)$, Eq.~\eqref{eq:bound1_no_alpha} gives an upper bound
	\begin{align}
		32B^2 \gamma^{-p}+ B\inf_{\rho\in(0,\gamma)} \left\{ 4\rho n + 12\sqrt{n}\int_{\rho}^\gamma \delta^{-p/2}d\delta \right\}
	\end{align}
	We choose $\gamma = n^{-1/(p+2)}$ and $\rho=n^{-1}$:
	{\small
	\begin{align*}
		32B^2 n^{\frac{p}{p+2}} + 4B + 12\sqrt{n}\left[ \left(\frac{2}{2-p}\right)\delta^{\frac{2-p}{2}}\right]_{n^{-1}}^{n^{-\frac{1}{p+2}}}
		&\leq 4B+ \left(32B^2  + 12B \left(\frac{2}{2-p}\right) \right)n^{\frac{p}{p+2}}
	\end{align*}}
	For the case $p=2$, we gain an extra factor of $\log(n)$ since the integral of $\delta^{-1}$ is the logarithm. For the parametric case \eqref{eq:vc_growth}, we choose $\gamma=n^{-1/2}$ and $\rho=n^{-1}$. Then Eq.~\eqref{eq:bound1_no_alpha} yields (for $n>8$),
	\begin{align*}
		V_n^0 &\leq 16B^2 d\log n + 4B + 12\sqrt{n}\int_{n^{-1}}^{n^{-1/2}} \sqrt{d\log(1/\delta)}d\delta \leq 16B^2 d\log n + 4B + 12\sqrt{d\log(n)}~.
	\end{align*}
	In the finite case, $\log \cN_2(\gamma, \F, n)\leq\log|\F|$ for any $\gamma$. We then have take $\gamma=0$ (one can see that this value is allowed for the particular case of a finite class; or, use a small enough value). Then,
	\begin{align*}
		V_n^0 &\leq 32B^2 \log |\F|~. 
	\end{align*}
Normalizing by $n$ yields the desired rates in the statement of the theorem. 
\end{proof}	

\begin{proof}[\textbf{Proof of Theorem~\ref{thm:lower}}]
	The first two lower bounds are proved in Lemma~\ref{lem:lower_p_greater_2} and \ref{lem:lower_p_less_2}. The lower bound for the parametric case follows from the i.i.d. lower bound in \citep{RakSriTsy14}.
\end{proof}

\begin{proof}[\textbf{Proof of Theorem~\ref{thm:optimistic}}]	
	For optimistic rates, we start with the upper bound in \eqref{eq:val_upper_alpha} and define $\G$ as above. We then appeal to Lemma~\ref{lem:reverse_dudley_with_alpha} and obtain 
	\begin{align}
		\label{eq:alpha_upper_bd_1}
		V_n^\alpha &\leq \alpha^{-1} 16 \log \cN_\infty(\gamma, \F,\z) + \alpha^{-1} \inf_{\rho\in(0,\gamma)} \left\{ 4\rho n + 16\log(\gamma/\rho) \int_{\rho}^\gamma \delta \log\cN_\infty(\delta, \F,\z) d\delta \right\} \ .
	\end{align}
For $\log \cN_\infty (\beta,\F,n)\leq \beta^{-p}$ decay of entropy for $p<2$, we take $\rho=(nB)^{-1}$, $\gamma=1$. The first term in \eqref{eq:alpha_upper_bd_1} can be taken to be zero, as we may take one function at scale $\gamma=1$.  The infimum in \eqref{eq:alpha_upper_bd_1} evaluates to
$$4 + 16\log(nB) \int_{1/(nB)}^1 \delta^{1-p}d\delta \leq 4+16\log(nB) \left[\frac{1}{2-p}\delta^{2-p}\right]_{1/(nB)}^1 \leq  4+16\log(nB)\frac{1}{2-p} \ .$$
For $p=2$, we gain an extra $\log(n)$ factor: $4+16(\log(nB))^2$.

For $p>2$, we take $\rho = n^{-\frac{1}{p-1}}$ and $\gamma=1$. Then infimum in \eqref{eq:alpha_upper_bd_1} evaluates to
\begin{align*}
	4n \cdot n^{-\frac{1}{p-1}}+16p^{-1}\log(n)\left[\frac{1}{2-p}\delta^{2-p}\right]_{n^{-\frac{1}{p-1}}}^1 \leq 4n^{\frac{p-2}{p-1}}+16p^{-1}\log(n)\frac{1}{2-p}n^{\frac{p-2}{p-1}} \ .
\end{align*}
For the parametric case \eqref{eq:vc_growth_infty}, we take $\gamma=1$ and $\rho = (nB)^{-1}$. Then \eqref{eq:alpha_upper_bd_1} is upper bounded by
\begin{align*}
	&4 + 16\log(nB) \int_{1/(nB)}^{1} d \delta \log(1/\delta) d\delta \leq 4 + 4d\log(nB) \ .
\end{align*}
The final optimistic rates are obtained by following the bound in \eqref{eq:generic_Lstar}.
\end{proof}

\subsection{Offset Rademacher Complexity and the Chaining Technique}
Let us recall the definition of sequential Rademacher complexity of a class $\F$
\begin{align}
	 \sup_{\x} \En \sup_{f\in\F}\left[ \sum_{t=1}^{n} \epsilon_t f(\x_t(\epsilon)) \right]
\end{align}
introduced in \citep{RakSriTew10}, where the expectation is over a sequence of independent Rademacher random variables $\epsilon= (\epsilon_1,\ldots,\epsilon_n)$ and the supremum is over all $\X$-valued trees of depth $n$. While this complexity both upper- and lower-bounds minimax regret for absolute loss, it fails to capture the possibly faster rates one can obtain for regression. We show below that modified, or \emph{offset}, versions of this complexity do in fact give optimal rates. These complexities have an extra quadratic term being subtracted off. Intuitively, this variance term ``extinguishes'' the $\sqrt{n}$-type fluctuations above a certain scale. Below this scale, complexity is given by the Dudley-type integral. The optimal balance of the scale gives the correct rates. As can be seen from the proof of Theorem~\ref{thm:upper}, the critical scale $\gamma$ is trivial (zero) for a finite case, then $n^{-1/2}$ for a parametric class, $n^{-1/(p+2)}$ for $p\in(0,2]$, and then becomes irrelevant (e.g. constant) at $p> 2$. Indeed, for $p>2$, the rate is given purely by sequential Rademacher complexity, as curvature of the loss does not help. In particular, can achieve these rates for $p>2$ by simply linearizing the square loss. The same phenomenon occurs in statistical learning with i.i.d. data \cite{RakSriTsy14}.

We remark that \cite{Mendelson14} studies bounds for estimation with squared loss for the empirical risk minimization procedure and observes that it is enough to only consider one-sided estimates rather than concentration statements. The offset sequential Rademacher complexities are of this one-sided nature. 

In Lemma~\ref{lem:expansion} below, we provide a bound on minimax regret via offset sequential Rademacher complexities.
\begin{lemma}
	\label{lem:expansion}
	The minimax value $V_n^\alpha$ of online regression with responses $y_t$ in a bounded interval $[-B,B]$ is upper bounded by 
	\begin{align}
		\label{eq:val_upper_alpha}
		V_n^\alpha &\leq \sup_{\x,\bmu,\bta} \En_\epsilon \sup_{f\in\F}\left[ \sum_{t=1}^{n} 4\epsilon_t \bta_t(\epsilon) (f(\x_t(\epsilon))-\bmu_t(\epsilon)) - (f(\x_t(\epsilon)) - \bmu_t(\epsilon))^2  -\alpha\bta_t(\epsilon)^2\right]
	\end{align}
	and
	\begin{align}
		\label{eq:val_upper_0_alpha}
		V_n^0 &\leq \sup_{\x,\bmu} \En_\epsilon \sup_{f\in\F}\left[ \sum_{t=1}^{n} 4B\epsilon_t  (f(\x_t(\epsilon))-\bmu_t(\epsilon)) - (f(\x_t(\epsilon)) - \bmu_t(\epsilon))^2  \right]
	\end{align}
	where $\x$ ranges over all $\X$-valued trees, $\bmu$ and $\bta$ over all $[-B,B]$-valued trees of depth $n$. Furthermore,
	\begin{align}
		\label{eq:lower_bd_rad_with_var}
		V_n^0 &\geq \sup_{\x,\bmu}\En \sup_{f\in\F}\left[ \sum_{t=1}^{n} B\epsilon_t (f(\x_t(\epsilon))-\boldsymbol{\mu}_t(\epsilon)) - (f(\x_t(\epsilon))-\boldsymbol{\mu}_t(\epsilon))^2  \right] 
	\end{align}
	where $\bmu$ ranges over $[-B/2, B/2]$-valued trees.
\end{lemma}

We now show that offset Rademacher complexities can be upper bounded by sequential entropies via the chaining technique. Lemma~\ref{lem:reverse_dudley} below is an analogue of the Dudley-type integral bound
\begin{align}
	\sup_{\x} \En \sup_{g\in\G}\left[ \sum_{t=1}^{n} \epsilon_t g(\x_t(\epsilon)) \right] \leq \inf_{\rho\in(0,1]} \left\{ 4\rho n + 12\sqrt{n}\int_{\rho}^1 \sqrt{\log\cN_2(\delta,\G,\z)}d\delta \right\} 
\end{align}
for sequential Rademacher proved in \citep{RakSriTew14}. Crucially, the upper bound of Lemma~\ref{lem:reverse_dudley} allows us to choose a critical scale $\gamma$.  
\begin{lemma}
	\label{lem:reverse_dudley}
	Let $\bta$ be a $[-B,B]$-valued tree of depth $n$. For any $\Z$-valued tree $\z$ and a class $\G$ of functions $\Z\to [-A,A]$ and any $\gamma\in(0,A]$,
	\begin{align*}
		\En \sup_{g\in\G}\left[ \sum_{t=1}^{n} 4 \epsilon_t \bta_t(\epsilon) g(\z_t(\epsilon)) - g(\z_t(\epsilon))^2 \right] &\leq 32B^2 \log \cN_2(\gamma, \G,\z) + B\inf_{\rho\in(0,\gamma)} \left\{ 4\rho n + 12\sqrt{n}\int_{\rho}^\gamma \sqrt{\log\cN_2(\delta,\G,\z)}d\delta \right\} 	
	\end{align*}
\end{lemma}
For optimistic rates, we can take advantage of an additional offset. This offset arises from the quadratic term due to the $\alpha$ multiple of the loss of the algorithm.
\begin{lemma}
	\label{lem:reverse_dudley_with_alpha}
	Let $\bta$ be a $[-B,B]$-valued tree of depth $n$. For any $\Z$-valued tree $\z$ and a class $\G$ of functions $\Z\to [-A,A]$, for any $\gamma\in(0,A]$,
	\begin{align}
		\label{eq:dudley_integral_with_alpha}
		\En \sup_{g\in\G}\left[ \sum_{t=1}^{n} 4 \epsilon_t \bta_t(\epsilon) g(\z_t(\epsilon)) - g(\z_t(\epsilon))^2 -\alpha\bta_t(\epsilon)^2 \right] &\leq \alpha^{-1} 16A^2 \log \cN_\infty(\gamma, \G,\z) \\
		&\hspace{-0cm}+ \alpha^{-1}\inf_{\rho\in(0,\gamma)} \left\{ 4\rho n + 16\log(\gamma/\rho) \int_{\rho}^\gamma \delta \log\cN_\infty(\delta, \G,\z) d\delta \right\}  	\notag
	\end{align}
\end{lemma}
The chaining arguments of Lemmas~\ref{lem:reverse_dudley} and \ref{lem:reverse_dudley_with_alpha}  are based on the following key finite-class lemma:
\begin{lemma}
	\label{lem:finite_lemma_with_etas}
	Let $\bta$ be a $[-B,B]$-valued tree of depth $n$. For a finite set $W$ of $[-A,A]$-valued trees of depth $n$, it holds that
\begin{align}
	\label{eq:finite_lemma_with_etas}
	\En \max_{\w\in W}\left[ \sum_{t=1}^{n} \epsilon_t \bta_t(\epsilon) \w_t(\epsilon) - C \w_t(\epsilon)^2 - \alpha\bta_t(\epsilon)^2  \right] \leq \min\left\{B^2(2C)^{-1}, A^2(2\alpha)^{-1}\right\}\log |W|
\end{align}
for any $C\geq 0$, $\alpha\geq 0$. It also holds that
\begin{align}
	\label{eq:finite_lemma_with_etas_no_var}
	\En \max_{\w\in W}\left[ \sum_{t=1}^{n} \epsilon_t \bta_t(\epsilon) \w_t(\epsilon) \right] \leq B\sqrt{2\log|W| \cdot \max_{\w\in W, \epsilon_{1:n}} \sum_{t=1}^n \w_n(\epsilon)^2} \ .
\end{align}
\end{lemma}

\begin{remark}
	\label{rem:vovk}
	Let us compare the upper bound of Lemma~\ref{lem:reverse_dudley} to the bound we may obtain via a metric entropy approach, as in the work of Vovk \cite{Vovk06metric}. Assume that $\F$ is a compact subset of $C(\X)$ equipped with supremum norm. The metric entropy, denoted by $\cH(\epsilon,\F)$, is the logarithm of the smallest $\epsilon$-net with respect to the sup norm on $\X$. An aggregating procedure over the elements of the net gives an upper bound (omitting constants and logarithmic factors) 
	\begin{align}
		\label{eq:balance_pollard}
		n\epsilon + \cH(\epsilon, \F)
	\end{align}
on regret \eqref{eq:standard_regret}. Here, $n\epsilon$ is the amount we lose from restricting the attention to the $\epsilon$-net, and the second term appears from aggregation over a finite set. While the balance \eqref{eq:balance_pollard} can yield correct rates for small classes, it fails to capture the optimal behavior for large nonparametric sets of functions. Indeed, for an $O(\epsilon^{-p})$ behavior of metric entropy, Vovk concludes the rate of $O\left(n^{\frac{p}{p+1}}\right)$. For $p\leq 2$, this is slower than the $O\left(n^{\frac{p}{p+2}}\right)$ rate one obtains from Lemma~\ref{lem:reverse_dudley} by trivially upper bounding the sequential entropy by metric entropy. The gain is due to the chaining technique, a phenomenon well-known in statistical learning theory. Our contribution is to introduce the same concepts to the domain of online learning. Let us also mention that sequential covering number of $\F$ is an ``empirical'' quantity and is finite even if we cannot upper bound metric entropy.
\end{remark}

\section{Further Examples}
For the sake of illustration we show bounds on minimax rates for a couple of examples.

\begin{example}[Sparse linear predictors]
Let $\G =\{g_1,\ldots,g_M\}$ be  a set of $M$ functions such that each $g_i : \X \mapsto [-1,1]$. Define $\F$ to be the convex combination of at most $s$ out of these $M$ functions. That is 
$$
\F = \left\{ \sum_{j=1}^s \alpha_j g_{\sigma_j} : \sigma_{1:s} \subset [M], \forall j, \alpha_j \ge 0, \sum_{j=1}^s \alpha_j = 1 \right\}
$$
For this example note that the sequential covering number can be easily upper bounded: we can choose $s$ out of $M$ functions in ${M \choose s}$ ways and further the $\ell_\infty$ metric entropy for convex combination of $s$ bounded functions at scale $\beta$ is bounded as $\beta^{-s}$. We conclude that
$$
\cN_2(\beta,\F,n) \le \left(\frac{eM}{s} \right)^s  \beta^{-s}
$$
From the main theorem, the upper bound is
$$
\tfrac{1}{n} V^0_n \le O\left(\frac{s \log(M/s) }{n}\right)
$$
\end{example}
\begin{example}[Besov Spaces]
	Let $\X$ be a compact subset of $\reals^d$. Let $\F$ be a ball in Besov space $B_{p,q}^s(\X)$. When $s>d/p$, pointwise metric entropy bounds at scale $\beta$ scale as $\Omega(\beta^{-d/s})$ \cite[p. 20]{Vovk06metric}. On the other hand, when $s\in(d/p,\infty)$, one can show that the space is a Banach space that is $p$-uniformly convex. From \citep{RakSriTew14}, it can be shown that sequential Rademacher can be upper bounded by $O(n^{1-1/p})$, yielding an bound on sequential entropy at scale $\beta$ as $O(\beta^{-p})$. These two controls together give the bound on the minimax rate. The generic forecaster with Rademacher complexity as relaxation (see Section~\ref{sec:relaxations}), enjoys the best of both of these rates. More specifically, we may identify the following regimes:
	\begin{itemize}
		\item If $s\geq d/2$, the minimax rate is $O\left(n^{\frac{2s}{2s+d}}\right)$.
		\item If $s< d/2$, the minimax rate depends on the interaction of $p$ and $d,s$:
		\begin{itemize}
			\item if $p> 1+\frac{d}{2s}$, the minimax rate is $O\left(n^{\frac{2s}{2s+d}}\right)$, as above.
			\item otherwise, the minimax rate is $O\left(n^{1-\frac{1}{p}}\right)$
		\end{itemize}
	\end{itemize}
\end{example}

\section{Lower Bounds}

The lower bounds will involve a notion of a ``dimension'' of $\F$ called the sequential fat-shattering dimension. Let us introduce this notion.

\begin{definition}
	An $\X$-valued tree of depth $d$ is said to be $\beta$-shattered by $\F$ if there exists an $\reals$-valued tree $\s$ of depth $d$ such that
		$$\forall \epsilon\in\{\pm1\}^d, ~\exists f^\epsilon \in\F ~~~\mbox{s.t.}~~~ \epsilon_t (f^\epsilon(\x_t(\epsilon))-\s_t(\epsilon))\geq \beta/2 $$
		for all $t\in\{1,\ldots,d\}$. The tree $\s$ is called a \emph{witness}. The largest $d$ for which there exists a $\beta$-shattered $\X$-valued tree is called the (sequential) fat-shattering dimension, denoted by $\fat_\beta(\F)$.
\end{definition}

The sequential fat-shattering dimension is related to sequential covering numbers as follows:
\begin{theorem}[\cite{RakSriTew14}] Let $\F$ be a class of functions $\X\to[-1,1]$. For any $\beta>0$, 
	$$\cN_2(\beta, \F, n)\leq \cN_\infty(\beta, \F, n) \leq \left( \frac{2en}{\beta}\right)^{\fat_\beta(\F)} \ .$$
\end{theorem}
Therefore, if $\log \cN_2(\beta, \F, n) \geq (c/\beta)^p$, then 
$$\fat_\beta(\F)\geq (c/\beta)^p/(\log(2en/\beta))\ .$$
The lower bounds will now be obtained assuming $\fat_\beta(\F) \geq C/\beta^p$ behavior of the fat-shattering dimension, and the resulting statement of Theorem~\ref{thm:lower} in terms of the sequential entropy growth will involve extra logarithmic factors, hidden in the $\tilde{\Omega}(\cdot)$ notation.

\begin{lemma}
	\label{lem:lower_p_greater_2}
	Consider the problem of online regression with responses bounded by $B=4$. For any class $\F$ of functions  $\X\to \left[-1,1\right]$ and any $\beta>0$ and $n=\fat_\beta(\F)$,
	$$\frac{1}{n}V_n^0 \geq \beta$$
	In particular, if $\fat_\beta(\F) \geq C/\beta^p$ for $p>0$, we have
	$$\frac{1}{n} V_n^0 \geq Cn^{-1/p} \ .$$
\end{lemma}

\begin{lemma}
	\label{lem:lower_p_less_2}
	For any class $\F'$ and $\beta>0$, there exists a modified class $\F$ such that $\fat_\beta(\F)\leq 2\fat_\beta(\F')+4$ and for $n>\fat_{\beta}(\F)$, 
	$$\frac{1}{n} V^0_n \geq C \left(2\sqrt{2} \beta \sqrt{\frac{\fat_\beta(\F)}{n}} - \beta^2\right) \ .$$
	In particular, when $p\in (0,2]$ and $\fat_\beta(\F)=C/\beta^p$,  
	$$\frac{1}{n} V^0_n \geq  C n^{-\frac{2}{p+2}} \ .$$
\end{lemma}


\section{Relaxations and Algorithms}
\label{sec:relaxations}
To design generic forecasters for the problem of online non-parametric regression we follow the recipe provided in  \citep{rakhlin2012relax}. It was shown in that paper that if one can find a relaxation $\mbf{Rel}_n$ (a sequence of mappings from observed data to reals) that satisfies initial and admissibility conditions then one can build estimators based on such relaxations. Specifically, we look for relaxations that satisfy the following initial condition 
\begin{align*}
\Rel{n}{x_{1:n},y_{1:n}} \ge - \inf_{f \in \F} \sum_{t=1}^n (f(x_t) - y_t)^2
\end{align*}
and the recursive admissibility condition that for any $t \in [n]$ and any $x_t \in \X$
\begin{align}\label{eq:admissibility}
\inf_{\hat{y}_t \in [-B,B]} \sup_{y_t \in [-B,B]}\left\{ (\hat{y}_t - y_t)^2 + \Rel{n}{x_{1:t},y_{1:t}} \right\}  \le \Rel{n}{x_{1:t-1},y_{1:t-1}} 
\end{align}
If a relaxation $\mbf{Rel}_n$ satisfies these two conditions then one can define an algorithm via 
$$
\hat{y}_t =  \argmin{\hat{y} \in [-B,B]} \sup_{y_t \in [-B,B]}\left\{ (\hat{y} - y_t)^2 + \Rel{n}{x_{1:t},y_{1:t}} \right\} 
$$
and for this forecast the associated bound on regret is automatically bounded as (see \citep{rakhlin2012relax} for details) :
$$
\Reg_n \le \Rel{n}{\cdot}
$$ 
Now further note that if $(\hat{y} - y_t)^2 + \Rel{n}{x_{1:t},(y_{1:t-1},y_t)}$ is a convex function of $y_t$ then the prediction takes a very simple form, as the supremum over $y_t$ is attained either at $B$ or $-B$. The prediction can be written as
\begin{align*}
\hat{y}_t & = \argmin{\hat{y} \in [-B,B]} \max\left\{(\hat{y} - B)^2 + \Rel{n}{x_{1:t},(y_{1:t-1},B)}, (\hat{y} + B)^2 + \Rel{n}{x_{1:t},(y_{1:t-1},-B)}\right\}
\end{align*}
Observe that the first term decreases as $\hat{y}$ increases to $B$ and likewise the second term monotonically decreases as $\hat{y}$ decreases to $-B$. Hence the solution to the above is given when both terms are equal (if this doesn't happen within the range $[-B,B]$ then we clip). In other words,
$$
\hat{y}_t = \mrm{Clip}\left(\frac{\Rel{n}{x_{1:t},(y_{1:t-1},B)} - \Rel{n}{x_{1:t},(y_{1:t-1},-B)} }{4B}\right)
$$ 
Hence, for any admissible relaxation such that $(\hat{y} - y_t)^2 + \Rel{n}{x_{1:t},(y_{1:t-1},y_t)}$ is a convex function of $y_t$, the above prediction based on the relaxation enjoys the bound on regret $\tfrac{1}{n} \mbf{Rel}_n$.

We now claim that the following conditional version of Equation \eqref{eq:val_upper_0_alpha} gives an admissible relaxation and leads to a method that enjoys the regret bounds shown in the first part of this paper.

\begin{lemma}\label{lem:condradrel}
The following relaxation is admissible :
\begin{align*}
\Rad_n(x_{1:t},y_{1:t}) = \sup_{\x,\bmu}\En_\epsilon \sup_{f\in\F}\left[ \sum_{j=t+1}^{n} 4 B\epsilon_j (f(\x_j(\epsilon))-\boldsymbol{\mu}_j(\epsilon)) - (f(\x_j(\epsilon))-\boldsymbol{\mu}_j(\epsilon))^2 - \sum_{j=1}^t (f(x_j) - y_j)^2  \right] 
\end{align*}
The forecast corresponding to this relaxation is given by 
$$
\hat{y}_t = \frac{\Rad_{n}(x_{1:t},(y_{1:t-1},B)) - \Rad_{n}(x_{1:t},(y_{1:t-1},-B)) }{4B}
$$
The above algorithm enjoys the regret bound of an offset Rademacher complexity:
$$
\Reg_n \le \sup_{\x,\bmu} \En_\epsilon \sup_{f\in\F}\left[ \sum_{t=1}^{n} 4B\epsilon_t  (f(\x_t(\epsilon))-\bmu_t(\epsilon)) - (f(\x_t(\epsilon)) - \bmu_t(\epsilon))^2  \right]
$$
\end{lemma}

Notice that since the regret bound for the above prediction based on the sequential Rademacher relaxation is exactly the one given in Equation \eqref{eq:val_upper_0_alpha}, the upper bounds provided for $V_n^0$ in Theorem \ref{thm:upper} also hold for the above algorithm.

\subsection{Recipe for designing online regression algorithms}
We now provide a schema for deriving forecasters for general online non-parametric regression problems:

\begin{enumerate}
\item Find relaxation $\mbf{Rel}_n$ such that 
$$
\Rad_n\left(x_{1:t},y_{1:t} \right) \le \Rel{n}{x_{1:t},y_{1:t}}
$$
and s.t. $(\hat{y} - y_t)^2 + \Rad_{n}\left(x_{1:t},(y_{1:t-1},y_t)\right)$ is a convex function of $y_t$
\item Check the condition 
$$
\sup_{x_t \in \X, p_t \in \Delta([-B,B])}\left\{  \Es{y_t \sim p_t}{\left(\Es{y_t \sim p_t}{y_t} - y_t\right)^2} + \Es{y_t \sim p_t}{\Rel{n}{x_{1:t},y_{1:t}}} \right\} \le \Rel{n}{x_{1:t-1},y_{1:t-1}} 
$$
\item Given $x_t$ on round $t$, the prediction $\hat{y}_t$ is given by 
$$\hat{y}_t = \mrm{Clip}\left(\frac{\Rel{n}{x_{1:t},(y_{1:t-1},B)} - \Rel{n}{x_{1:t},(y_{1:t-1},-B)} }{4B}\right)$$
\end{enumerate}

\begin{proposition}\label{prop:recipe}
Any algorithm derived from the above schema using relaxation $\mbf{Rel}_n$ enjoys a bound  
$$\Reg_n \le \tfrac{1}{n} \Rel{n}{\cdot}$$
on regret.
\end{proposition}

\noindent{\bf Example : Finite class of experts}\\
As an example of estimator derived from the schema we first consider the simple case $|\F| < \infty$. 

\begin{corollary}\label{cor:exp}
The following is an admissible relaxation :
$$
\Rel{n}{x_{1:t},y_{1:t}} = B^2 \log\left(\sum_{f\in\F} \exp\left(  - B^{-2} \sum_{j=1}^t (f(x_j) - y_j)^2\right) \right) 
$$
It leads to the following algorithm
\begin{align*}
\hat{y}_t & =  \mrm{Clip}\left(\frac{B}{4} \log\left(\frac{\sum_{f\in\F} \exp\left(  - B^{-2} \sum_{j=1}^{t-1} (f(x_j) - y_j)^2 - B^{-2} (f(x_t) - B)^2\right)}{\sum_{f\in\F} \exp\left(  - B^{-2} \sum_{j=1}^{t-1} (f(x_j) - y_j)^2  - B^{-2} (f(x_t) + B)^2\right)} \right)  \right)
\end{align*}
and enjoys a regret bound
$
\Reg_n \le B^2 \log\left| \F \right| \ .
$
\end{corollary}

\noindent{\bf Example : Linear regression}\\
Next, consider the problem of online linear regression in $\reals^d$. Here $\F$ is the class of linear functions. For this problem we consider a slightly modified notion of regret :
$$
\sum_{t=1}^n (\hat{y}_t - y_t)^2 - \inf_{f \in \F}\left\{ \sum_{t=1}^n (f^\top x_t - y_t)^2 + \lambda \norm{f}_2^2  \right\}
$$
This regret can be seen alternatively as regret if we assume that on rounds $-d+1$ to $0$ Nature plays  $(\lambda e_1,0)$,\ldots, $(\lambda e_d,0)$, where $\{e_i\}$ are the standard basis vectors, and that on these rounds the learner (knowing this) predicts $0$, thus incurring zero loss over these initial rounds. Hence we can readily apply the schema for designing an algorithm for this problem.

\begin{corollary}
	\label{cor:linreg}
For any $\lambda>0$, the following is an admissible relaxation
\begin{align*}
\Rel{n}{x_{1:t},y_{1:t}} &= \norm{\sum_{j=1}^t y_j z_j}^2_{\left(\sum_{j=1}^t z_j z_j^\top + \lambda I\right)^{-1}} + 4 B^2 \log\left(\frac{\left(\frac{n}{d}\right)^d}{\Delta\left(\sum_{j=1}^t z_j z_j^\top + \lambda I\right)}\right)  - \sum_{j=1}^t y_j^2 \ .
\end{align*}
It leads to the Vovk-Azoury-Warmuth forecaster \citep{Vovk98,AzouryWarmuth01}: 
\begin{align*}
\hat{y}_t & = \mrm{Clip}\left( x_t^\top \left(\sum_{j=1}^t x_j x_j^\top + \lambda I\right)^{-1}  \left(\sum_{j=1}^{t-1} y_j x_j\right)
\right)
\end{align*}
and enjoys the following upper bound on regret:
\begin{align*}
\frac{1}{n}\sum_{t=1}^n (\hat{y}_t - y_t)^2 \le \frac{1}{n}  \sum_{t=1}^n (f^\top x_t - y_t)^2 + \frac{\lambda}{2 n} \norm{f}_2^2 + \frac{4 d B^2 \log\left(\frac{n}{\lambda d}\right)}{n}
\end{align*}
\end{corollary}

\bibliographystyle{plain}
\bibliography{paper}
\appendix

\newpage
\section{Proofs}

\begin{proof}[\textbf{Proof of Lemma~\ref{lem:expansion}}]
Let us now study the value \eqref{eq:def_value}. We will do so ``from inside out'' by considering the last step $t=n$, then working our way back to $t=1$. Given a value $x_n$, by the minimax theorem,
\begin{align}
	\label{eq:one_step_quadratic}
	&\inf_{q_n}\sup_{p_n}\En_{\hat{y}_n\sim q_n,y_n\sim p_n} \left\{ (1-\alpha)(\hat{y}_n-y_n)^2 + \sup_{f\in\F}\sum_{t=1}^n -(f(x_t)-y_t)^2 \right\} \\
	&= \sup_{p_n}  \left\{ (1-\alpha)\inf_{\hat{y}_n}\En_{y_n}(\hat{y}_n-y_n)^2 + \En_{y_n}\sup_{f\in\F}\sum_{t=1}^n -(f(x_t)-y_t)^2 \right\} \notag\\
	&= \sup_{p_n}  \En_{y_n}\left\{ (1-\alpha)(\mu_n-y_n)^2 + \sup_{f\in\F}\sum_{t=1}^n -(f(x_t)-y_t)^2 \right\} 
\end{align}
where $\mu_n = \En[y_n]$ under the distribution $p_n$ with support on $[-B,B]$. Observe that
\begin{align}
	\label{eq:diff_of_squares}
	(\mu_n-y_n)^2 - (f(x_n)-y_n)^2 = 2(y_n-\mu_n)(f(x_n)-\mu_n) - (f(x_n)-\mu_n)^2
\end{align}
and hence the expression in \eqref{eq:one_step_quadratic} can be written as
\begin{align*}
	\sup_{p_n} \En_{y_n}\sup_{f\in\F}\left[ \sum_{t=1}^{n-1} -(f(x_t)-y_t)^2  + \left\{ 2(y_n-\mu_n)(f(x_n)-\mu_n) - (f(x_n) - \mu_n)^2 -\alpha(\mu_n-y_n)^2\right\}\right] 
\end{align*}
Continuing in this fashion back to $t=1$, the minimax value is equal to 
\begin{align}
	\label{eq:unrolled_value}
	V_n^\alpha &=\multiminimax{\sup_{x_t}\sup_{p_t}\En_{y_t}}_{t=1}^n \left\{ \sup_{f\in\F}\left[ \sum_{t=1}^{n} 2(y_t-\mu_t)(f(x_t)-\mu_t) - (f(x_t) - \mu_t)^2  -\alpha(\mu_t-y_t)^2\right] \right\}.
\end{align}
The supremum over $p_t$ can now be upper bounded by the supremum over the mean $\mu_t\in[-B,B]$ and a zero-mean distribution $p_t'$ with support on $[-B,B]$. Denoting by $\eta_t$ a random variable with this distribution $p_t'$, the variable $\mu_t+\eta_t$ is then in $[-2B,2B]$. We upper bound \eqref{eq:unrolled_value} by
\begin{align}
	\label{eq:unrolled_value_with_eta}
	V_n^\alpha &\leq \multiminimax{\sup_{x_t}\sup_{p_t',\mu_t}\En_{\eta_t}}_{t=1}^n \left\{ \sup_{f\in\F}\left[ \sum_{t=1}^{n} 2\eta_t(f(x_t)-\mu_t) - (f(x_t) - \mu_t)^2  -\alpha\eta_t^2\right] \right\}.
\end{align}
Since the $-\alpha\eta^2$ term does not depend on $f$, we use linearity of expectation to write
\begin{align}
	V_n^\alpha &= \multiminimax{\sup_{x_t}\sup_{p_t',\mu_t}\En_{\eta_t}}_{t=1}^n \left\{ \sup_{f\in\F}\left[ \sum_{t=1}^{n} 2\eta_t(f(x_t)-\mu_t) - (f(x_t) - \mu_t)^2  - \mathcal{D}(p'_1,\ldots,p'_n) \right] \right\}
\end{align}
where 
$$\mathcal{D}(p'_1,\ldots,p'_n) = \frac{1}{n}\sum_{t=1}^{n} \alpha\En\eta_t^2.$$
We now symmetrize the linear term. Let $(\eta'_t)$ be a sequence tangent to $(\eta_t)$ (that is, $\eta_t$ and $\eta_t'$ are i.i.d. conditionally on $\eta_{1:t-1}$). We write $\mu_t = \En[\eta'_t]$ and use convexity of the supremum to arrive at an upper bound
\begin{align}
	V_n^\alpha &\leq \multiminimax{\sup_{x_t}\sup_{p_t',\mu_t}\En_{\eta_t}}_{t=1}^n \left\{ \sup_{f\in\F}\left[ \sum_{t=1}^{n} 2(\eta_t-\eta'_t)(f(x_t)-\mu_t) - (f(x_t) - \mu_t)^2  - \mathcal{D}(p'_1,\ldots,p'_n) \right]\right\}  \label{eq:symmetrized_subg_no_eps}\\
	&= \multiminimax{\sup_{x_t}\sup_{p_t',\mu_t}\En_{\eta_t,\eta'_t}\Ex_{\epsilon_t}}_{t=1}^n \left\{ \sup_{f\in\F}\left[ \sum_{t=1}^{n} 2\epsilon_t(\eta_t-\eta'_t)(f(x_t)-\mu_t) - (f(x_t) - \mu_t)^2  -\mathcal{D}(p'_1,\ldots,p'_n)\right] \right\}  
\end{align}
where in the second equality holds because $\eta'_t$ and $\eta_t$ are i.i.d. from $p'_t$, conditionally on the past observations. We now split the above supremum over $f$ into two parts, thus passing to the upper bound
\begin{align*}
	&\multiminimax{\sup_{x_t}\sup_{p_t,\mu_t}\Ex_{\eta_t,\eta'_t}\Ex_{\epsilon_t}}_{t=1}^n \left\{ \sup_{f\in\F}\left[ \sum_{t=1}^{n} 2\epsilon_t \eta_t (f(x_t)-\mu_t) - \frac{1}{2}(f(x_t) - \mu_t)^2  -\frac{1}{2}\mathcal{D}(p'_1,\ldots,p'_n)\right] \right\}  \\
	&+ \multiminimax{\sup_{x_t}\sup_{p_t,\mu_t}\Ex_{\eta_t,\eta'_t}\Ex_{\epsilon_t}}_{t=1}^n \left\{ \sup_{f\in\F}\left[ \sum_{t=1}^{n} -2\epsilon_t \eta'_t(f(x_t)-\mu_t) - \frac{1}{2}(f(x_t) - \mu_t)^2  -\frac{1}{2}\mathcal{D}(p'_1,\ldots,p'_n)\right] \right\} \\
	&= \multiminimax{\sup_{x_t}\sup_{p'_t,\mu_t}\Ex_{\eta_t \sim p_t}\Ex_{\epsilon_t}}_{t=1}^n \left\{ \sup_{f\in\F}\left[ \sum_{t=1}^{n} 4\epsilon_t \eta_t (f(x_t)-\mu_t) - (f(x_t) - \mu_t)^2  -\mathcal{D}(p'_1,\ldots,p'_n)\right] \right\} \\
	&= \multiminimax{\sup_{x_t}\sup_{p'_t,\mu_t}\Ex_{\eta_t \sim p_t}\Ex_{\epsilon_t}}_{t=1}^n \left\{ \sup_{f\in\F}\left[ \sum_{t=1}^{n} 4\epsilon_t \eta_t (f(x_t)-\mu_t) - (f(x_t) - \mu_t)^2  -\alpha\eta_t^2\right] \right\} \\
	&\leq \multiminimax{\sup_{x_t}\sup_{\mu_t,\eta_t}\Ex_{\epsilon_t}}_{t=1}^n \left\{ \sup_{f\in\F}\left[ \sum_{t=1}^{n} 4\epsilon_t \eta_t (f(x_t)-\mu_t) - (f(x_t) - \mu_t)^2  -\alpha\eta_t^2\right] \right\} \\
	&= \sup_{\x,\bmu,\bta} \En_\epsilon \sup_{f\in\F}\left[ \sum_{t=1}^{n} 4\epsilon_t \bta_t(\epsilon) (f(\x_t(\epsilon))-\bmu_t(\epsilon)) - (f(\x_t(\epsilon)) - \bmu_t(\epsilon))^2  -\alpha\bta_t(\epsilon)^2\right]
\end{align*}
This proves the first statement. For the case $\alpha=0$, we have
\begin{align*}
	V_n^0 &\leq \sup_{\x,\bmu,\bta} \En_\epsilon \sup_{f\in\F}\left[ \sum_{t=1}^{n} 4\epsilon_t \bta_t(\epsilon) (f(\x_t(\epsilon))-\bmu_t(\epsilon)) - (f(\x_t(\epsilon)) - \bmu_t(\epsilon))^2 \right] \\
	&= \multiminimax{\sup_{x_t,\mu_t,\eta_t}\Ex_{\epsilon_t}}_{t=1}^n \left\{\sup_{f\in\F}\left[ \sum_{t=1}^{n} 4\epsilon_t \eta_t (f(x_t)-\mu_t) - (f(x_t) - \mu_t(\epsilon))^2 \right] \right\}
\end{align*}
Since each $\eta_t$ range over $[-B,B]$, we can represent it as $B$ times the expectation of a  random variable $u_t\in\{-1,1\}$. Denoting this distribution by $q_t$, by Jensen's inequality 
\begin{align*}
	V_n^0 &\leq \multiminimax{\sup_{x_t,\mu_t,q_t}\Ex_{\epsilon_t}}_{t=1}^n \left\{\sup_{f\in\F}\left[ \sum_{t=1}^{n} 4\epsilon_t \En(u_t)B(f(x_t)-\mu_t) - (f(x_t) - \mu_t(\epsilon))^2 \right] \right\} \\
	&\leq \multiminimax{\sup_{x_t,\mu_t,q_t}\Ex_{u_t}\Ex_{\epsilon_t}}_{t=1}^n \left\{\sup_{f\in\F}\left[ \sum_{t=1}^{n} 4\epsilon_t u_t B(f(x_t)-\mu_t) - (f(x_t) - \mu_t(\epsilon))^2 \right] \right\} \\
	&= \multiminimax{\sup_{x_t,\mu_t,u_t}\Ex_{\epsilon_t}}_{t=1}^n \left\{\sup_{f\in\F}\left[ \sum_{t=1}^{n} 4\epsilon_t u_t B(f(x_t)-\mu_t) - (f(x_t) - \mu_t(\epsilon))^2 \right] \right\} \\
	&= \multiminimax{\sup_{x_t,\mu_t}\Ex_{\epsilon_t}}_{t=1}^n \left\{\sup_{f\in\F}\left[ \sum_{t=1}^{n} 4\epsilon_t B(f(x_t)-\mu_t) - (f(x_t) - \mu_t(\epsilon))^2 \right] \right\} 
\end{align*}
which is the same as the desired upper bound in \eqref{eq:val_upper_0_alpha}, in the tree notation.

As for the lower bound,
	Recall from Eq.~\eqref{eq:unrolled_value} that the value with $\alpha=0$ is equal to 
	\begin{align}
		\label{eq:unrolled_value_alpha_0}
		V_n^0 &=\multiminimax{\sup_{x_t}\sup_{p_t}\En_{y_t}}_{t=1}^n \left\{ \sup_{f\in\F}\left[ \sum_{t=1}^{n} 2(y_t-\mu_t)(f(x_t)-\mu_t) - (f(x_t) - \mu_t)^2 \right] \right\}.
	\end{align} 
	For the purposes of a lower bound, let us pick particular distributions $p_t$ as follows. Let $\epsilon_1,\ldots,\epsilon_n$ be independent Rademacher random variables. Fix a $[-B/2,B/2]$-valued tree $\bmu$. Let $y_t=\bmu_t(\epsilon_{1:t-1})+(B/2)\epsilon_{t}$. Hence, $y_t\in[-B,B]$ as required. We can then lower bound the above expression as
	\begin{align*}
		V_n^0 &\geq \sup_{\bmu}\multiminimax{\sup_{x_t}\En_{\epsilon_t}}_{t=1}^n \left\{ \sup_{f\in\F}\left[ \sum_{t=1}^{n} 2\epsilon_t (f(x_t)-\bmu_t(\epsilon)) - (f(x_t)-\bmu_t(\epsilon))^2  \right] \right\} \\
		&= \sup_{\x,\bmu}\En \sup_{f\in\F}\left[ \sum_{t=1}^{n} B\epsilon_t (f(\x_t(\epsilon))-\boldsymbol{\mu}_t(\epsilon)) - (f(\x_t(\epsilon))-\boldsymbol{\mu}_t(\epsilon))^2  \right] 
	\end{align*}
\end{proof}

\begin{proof}[\textbf{Proof of Lemma~\ref{lem:finite_lemma_with_etas}}]
	For any $\lambda>0$,
	\begin{align*}
		&\En \max_{\w\in W}\left[ \sum_{t=1}^{n} \epsilon_t \bta_t(\epsilon) \w_t(\epsilon) - C \w_t(\epsilon)^2 - \alpha\bta_t(\epsilon)^2  \right] \leq \frac{1}{\lambda} \log \En \sum_{\w\in W} \exp\left\{ \sum_{t=1}^{n} \lambda\epsilon_t \bta_t(\epsilon) \w_t(\epsilon) - \lambda C \w_t(\epsilon)^2 - \lambda\alpha \bta_t(\epsilon)^2  \right\}
	\end{align*}
Conditioning on $\epsilon_{1:n-1}$, we analyze
\begin{align}
	&\En \left[ \sum_{\w\in W} \exp\left\{ \sum_{t=1}^{n} \lambda\epsilon_t \bta_t(\epsilon) \w_t(\epsilon) - \lambda C \w_t(\epsilon)^2 - \lambda\alpha \bta_t(\epsilon)^2  \right\} ~\middle|~ \epsilon_{1:n-1} \right] \notag \\
	&=  \sum_{\w\in W} \exp\left\{ \sum_{t=1}^{n-1} \lambda \epsilon_t \bta_t(\epsilon) \w_t(\epsilon))- \sum_{t=1}^{n} \lambda C\w_t(\epsilon)^2 - \sum_{t=1}^{n} \lambda\alpha \bta_t(\epsilon)^2 \right\} \En \left[ \exp\left\{  \lambda\epsilon_n \bta_n(\epsilon) \w_n(\epsilon)\right\} ~\middle|~ \epsilon_{1:n-1} \right] \notag\\
	&\leq \sum_{\w\in W} \exp\left\{ \sum_{t=1}^{n-1} \lambda \epsilon_t \eta_t \w_t(\epsilon)- \sum_{t=1}^{n-1} \lambda C\w_t(\epsilon)^2 - \sum_{t=1}^{n-1} \lambda\alpha \bta_t(\epsilon)^2 \right\} \exp\left\{ \lambda^2 \bta_n(\epsilon)^2 \w_n(\epsilon)^2/2 - \lambda C\w_n(\epsilon)^2 - \lambda\alpha \bta_n(\epsilon)^2\right\} \label{eq:one_term_peel_off} 
\end{align}
The choice $\lambda=2C/B^2$ ensures
$$\lambda^2 \bta_n(\epsilon)^2 \w_n(\epsilon)^2/2 - \lambda C\w_n(\epsilon)^2\leq 0$$
Alternatively, the choice $\lambda = 2\alpha/A^2$ ensures
$$\lambda^2 \bta_n(\epsilon)^2 \w_n(\epsilon)^2/2 - \lambda\alpha \bta_n(\epsilon)^2 \leq 0$$
In both cases, the exponential factor peeled off in \eqref{eq:one_term_peel_off} is no greater than $1$.
We proceed all the way to $t=1$ to arrive at an upper bound of 
$$\frac{1}{\lambda}\log \sum_{\w\in W} \exp\{0\} = \min\left\{B^2(2C)^{-1}, A^2(2\alpha)^{-1}\right\}\log |W| \ .$$
The second statement (which already appears in \cite{RakSriTew10}) is proved similarly, except the tuning value $\lambda$ is chosen at the end, and we need to account for the worst-case $\ell_2$ norm along any paths. 
For any tree $\w\in W$,
\begin{align*}
	\En \left[ \exp\left\{ \sum_{t=1}^{n} \lambda\epsilon_t \bta_t(\epsilon) \w_t(\epsilon) \right\} ~\middle|~ \epsilon_{1:n-1} \right] &\leq \exp\left\{ \sum_{t=1}^{n-1} \lambda\epsilon_t \bta_t(\epsilon) \w_t(\epsilon) \right\} \exp\left\{  B^2 \lambda^2 \w_n(\epsilon)^2/2\right\}  \\
	&\leq \exp\left\{ \sum_{t=1}^{n-1} \lambda\epsilon_t \bta_t(\epsilon) \w_t(\epsilon) \right\} \max_{\epsilon_n}\exp\left\{  B^2 \lambda^2 \w_n(\epsilon)^2/2\right\}
\end{align*}
Continuing in this fashion backwards to $t=1$, for any $\w\in W$
$$\En \left[ \exp\left\{ \sum_{t=1}^{n} \lambda\epsilon_t \bta_t(\epsilon) \w_t(\epsilon) \right\} \right] \leq \max_{\epsilon_1,\ldots,\epsilon_n} \exp\left\{  B^2 (\lambda^2/2) \sum_{t=1}^n \w_n(\epsilon)^2 \right\} $$
and thus
$$\En \left[ \sum_{\w\in W}\exp\left\{ \sum_{t=1}^{n} \lambda\epsilon_t \bta_t(\epsilon) \w_t(\epsilon) \right\} \right] \leq |W|\max_{\epsilon_1,\ldots,\epsilon_n}\max_{\w\in W} \exp\left\{  B^2 (\lambda^2/2) \sum_{t=1}^n \w_n(\epsilon)^2 \right\} \ . $$
Choosing 
$$\lambda = \sqrt{\frac{2\log |W|}{B^2 \max_{\epsilon_{1:n}, \w\in W} \sum_{t=1}^n \w_n(\epsilon)^2}}$$
we obtain
\begin{align*}
\En \max_{\w\in W}\left[ \sum_{t=1}^{n} \epsilon_t \bta_t(\epsilon) \w_t(\epsilon) \right] &\leq \frac{1}{\lambda}\log \En \left[ \sum_{\w\in W}\exp\left\{ \sum_{t=1}^{n} \lambda\epsilon_t \bta_t(\epsilon) \w_t(\epsilon) \right\} \right] \leq B\sqrt{2\log|W| \cdot \max_{\w\in W, \epsilon_{1:n}} \sum_{t=1}^n \w_n(\epsilon)^2}
\end{align*}
\end{proof}

\begin{proof}[\textbf{Proof of Lemma~\ref{lem:reverse_dudley}}]
Let $V'$ be a sequential $\gamma$-cover of $\G$ on $\z$ in the $\ell_2$ sense, i.e.
$$\forall \epsilon,~~ \forall g\in\G,~~ \exists \v\in V' \mbox{~~~s.t.~~~} \frac{1}{n}\sum_{t=1}^n (g(\z_t(\epsilon))-\v_t(\epsilon))^2 \leq \gamma^2$$
Let us augment $V'$ to include the all-zero tree, and denote the resulting set by $V=V'\cup \{\boldsymbol{0}\}$. Denote by $\v[\epsilon,g]$ a $\gamma$-close tree promised above, but we leave the choice for later. Then for any $c\in[0,1]$
\begin{align}
	\label{eq:dudley_initial_split}
	&\En \sup_{g\in\G}\left[ \sum_{t=1}^{n} 4\epsilon_t \bta_t(\epsilon) g(\z_t(\epsilon)) - g(\z_t(\epsilon))^2  \right] \\
	&=\En \sup_{g\in\G}\left[ \sum_{t=1}^{n} 4\epsilon_t \bta_t(\epsilon) \Big(g(\z_t(\epsilon))-\v[\epsilon,g]_t(\epsilon)\Big) - \Big(g(\z_t(\epsilon))^2-c^2\v[\epsilon,g]_t(\epsilon)^2\Big) \right.\\ 	
	&\left.\hspace{1in} +\Big(4\epsilon_t\bta_t(\epsilon)\v[\epsilon,g]_t(\epsilon) - c^2\v[\epsilon,g]_t(\epsilon)^2  \Big)\right] \\
	&\leq \En \sup_{g\in\G}\left[ \sum_{t=1}^{n} 4\epsilon_t \bta_t(\epsilon)\Big(g(\z_t(\epsilon))-\v[\epsilon,g]_t(\epsilon)\Big) - \sum_{t=1}^{n} \Big(g(\z_t(\epsilon))^2-c^2\v[\epsilon,g]_t(\epsilon)^2\Big) \right]  \\
	&~~~~~+ \En \max_{\v\in V'}\left[\sum_{t=1}^n 4\epsilon_t\bta_t(\epsilon)\v_t(\epsilon) - c^2\v_t(\epsilon)^2 \right]
\end{align}
We now claim that for any $\epsilon,g$ there exists an element $\v[\epsilon,g]\in V$ such that
\begin{align}
	\label{eq:desired_l2_norm_relation}
	\sum_{t=1}^{n} g(\z_t(\epsilon))^2\geq c^2\sum_{t=1}^{n}\v[\epsilon,g]_t(\epsilon)^2 
\end{align}
and so we can drop the corresponding negative term in the supremum over $\G$. To prove this claim, first consider the easy case $\frac{1}{n}\sum_{t=1}^{n} g(\z_t(\epsilon))^2 \leq C^2\gamma^2$, where $C=\frac{c}{1-c}$. Then we may choose $\boldsymbol{0}\in V$ as a tree that provides a sequential $C\gamma$-cover in the $\ell_2$ sense. Clearly, \eqref{eq:desired_l2_norm_relation} is then satisfied with this choice of $\v[\epsilon,g]=\boldsymbol{0}$. Now, assume $\frac{1}{n}\sum_{t=1}^{n} g(\z_t(\epsilon))^2 > C^2\gamma^2$. Fix any tree $\v[\epsilon,g]\in V$ that is $\gamma$-close in the $\ell_2$ sense to $g$ on the path $\epsilon$. Denote $u=(\v[\epsilon,g]_1(\epsilon),\ldots, \v[\epsilon,g]_n(\epsilon))$ and $h=(g(\z_1(\epsilon)),\ldots, g(\z_n(\epsilon)))$. Thus, we have that $\|u-h\|\leq \gamma$ and $\|h\|\geq C\gamma$ for the norm $\|h\|^2=\frac{1}{n}\sum_{t=1}^n h_t^2$. 
Then $$\|u\|\leq \|u-h\|+\|h\| \leq \gamma + \|h\| \leq (C^{-1}+1) \|h\|$$
and thus $\|h\|\geq c\|u\|$ as desired. By choosing $c=1/2$, we have $C=1$ and thus the zero tree also provides a $\gamma$-cover.

We conclude that
\begin{align}
	\label{eq:dud_eq_split_no_square}
	\En \sup_{g\in\G}\left[ \sum_{t=1}^{n} 4\epsilon_t\bta_t(\epsilon) g(\z_t(\epsilon)) - g(\z_t(\epsilon))^2   \right] &\leq 4\En \sup_{g\in\G}\left[ \sum_{t=1}^{n} \epsilon_t \bta_t(\epsilon)\Big(g(\z_t(\epsilon))-\v[\epsilon,g]_t(\epsilon)\Big)  \right] \\ 
	& + \En \max_{\v\in V'}\left[\sum_{t=1}^{n}4\epsilon_t\bta_t(\epsilon)\v_t(\epsilon) - (1/4)\v_t(\epsilon)^2\right] \label{eq:term2}
\end{align}
where $\v[\epsilon,g]$ is defined to be the all-zero tree if $\frac{1}{n}\sum_{t=1}^{n} g(\z_t(\epsilon))^2 \leq \gamma^2$ and otherwise as an element of the cover $V'$ that is $\gamma$-close to $g$ on the path $\epsilon$.

By Lemma~\ref{lem:finite_lemma_with_etas}, the term \eqref{eq:term2} is upper bounded as
$$\En_\epsilon \max_{\v\in V'}\left[\sum_{t=1}^{n}4\epsilon_t\bta_t(\epsilon)\v_t(\epsilon) - (1/4)\v_t(\epsilon)^2 \right] \leq 32B^2 \log \cN_2(\gamma, \G,\z)$$
We now turn to the analysis of the first term on the right-hand side of \eqref{eq:dud_eq_split_no_square}. Let $\v[\epsilon,g]$ be denoted by $\v[\epsilon,g]^0$ and $V$ be denoted by $V^0$. Let $V^j$ denote a sequential $(2^{-j}\gamma)$-cover of $\G$ on the tree $\z$, for $j=1,\ldots,N$, $N\geq 1$ to be specified later. We can now write 
\begin{align*}
	&\En \sup_{g\in\G}\left[ \sum_{t=1}^{n} \epsilon_t \bta_t(\epsilon)\Big(g(\z_t(\epsilon))-\v[\epsilon,g]^0_t(\epsilon)\Big)  \right] \\
	&= \En \sup_{g\in\G}\left[ \sum_{t=1}^{n} \epsilon_t \bta_t(\epsilon) \Big(g(\z_t(\epsilon))-\v[\epsilon,g]^N_t(\epsilon)\Big) +  \sum_{t=1}^{n} \sum_{j=1}^N\epsilon_t \bta_t(\epsilon)\Big(\v[\epsilon,g]^j_t(\epsilon)-\v[\epsilon,g]^{j-1}_t(\epsilon)\Big) \right] \\
	&\leq \En \sup_{g\in\G}\left[ \sum_{t=1}^{n} \epsilon_t \bta_t(\epsilon)\Big(g(\z_t(\epsilon))-\v[\epsilon,g]^N_t(\epsilon)\Big) \right] + \sum_{j=1}^N \En \sup_{g\in\G}\left[ \sum_{t=1}^{n} \epsilon_t\bta_t(\epsilon) \Big(\v[\epsilon,g]^j_t(\epsilon)-\v[\epsilon,g]^{j-1}_t(\epsilon)\Big) \right]
\end{align*}
From here, the analysis is very similar to the one in \cite{RakSriTew14}, except for the additional random variables $\bta_t(\epsilon)$ multiplying the differences, and also for the minor fact that $\v[\epsilon,g]^0$ is defined as $\boldsymbol{0}$ for some $(g,\epsilon)$ pairs. This latter fact, however, does not affect the proof since $\boldsymbol{0}$ does provide a valid $\gamma$-cover when it is used. 

First, by Cauchy-Schwartz inequality,
\begin{align*}
	\En \sup_{g\in\G}\left[ \sum_{t=1}^{n} \epsilon_t \bta_t(\epsilon)\Big(g(\z_t(\epsilon))-\v[\epsilon,g]^N_t(\epsilon)\Big) \right] &\leq n \En \sup_{g\in\G}\left[ \sum_{t=1}^{n} \left( \frac{\epsilon_t \bta_t(\epsilon)}{\sqrt{n}} \right) \left(\frac{1}{\sqrt{n}}\Big(g(\z_t(\epsilon))-\v[\epsilon,g]^N_t(\epsilon)\Big)\right) \right] \\
	&\leq n \En \left(\frac{1}{n}\sum_{t=1}^n \bta_t(\epsilon)^2\right)^{-1/2} \beta_N  \\
	&\leq B \beta_N n
\end{align*}
where $\beta_j = 2^{-j}\gamma$. For the second term, fix a particular $j$ and consider all pairs $(\v^s,\v^r)$ with $\v^s\in V^j$ and $\v^r\in V^{j-1}$. For each such pair, define a tree $\w^{(s,r)}$ by
\begin{align*}
\w^{(s,r)}_t(\epsilon) = 
        \begin{cases} 
        \v^s_t(\epsilon)-\v^r_t(\epsilon) & \text{if there exists } g\in\G \mbox{ s.t. } \v^s = \v[g,\epsilon]^{j}, \v^r = \v[g, \epsilon]^{j-1} \\
        0 &\text{otherwise.}
        \end{cases}
\end{align*}
for all $t\in [n]$ and $\epsilon\in\{\pm1\}^n$. One can check that the tree is well-defined, and we set $$W_j = \left\{w^{(s,r)}: 1\leq s\leq |V_j|, 1\leq r\leq |V_{j-1}| \right\}.$$ Then for any $j\in[N]$ and $\epsilon$,
\begin{align*}
	\sup_{g\in\G}\left[ \sum_{t=1}^{n} \epsilon_t \bta_t(\epsilon)\Big(\v[\epsilon,g]^j_t(\epsilon)-\v[\epsilon,g]^{j-1}_t(\epsilon)\Big) \right] \leq \max_{\w\in W}\left[ \sum_{t=1}^{n} \epsilon_t \bta_t(\epsilon)\w_t(\epsilon) \right]
\end{align*}
By the argument outlined in \citep{RakSriTew14}, for any $\w\in W^j$ and any path $\epsilon$,
$$\sqrt{\sum_{t=1}^n \w_t(\epsilon)^2}\leq 3\sqrt{n}\beta_j \ .$$
Putting everything together, and using Lemma~\ref{lem:finite_lemma_with_etas}, 
\begin{align*}
	&\En_\epsilon \sup_{g\in\G}\left[ \sum_{t=1}^{n} \epsilon_t \bta_t(\epsilon)\Big(g(\z_t(\epsilon))-\v[\epsilon,g]^0_t(\epsilon)\Big)  \right] \leq B \beta_N n + B \sqrt{n}\sum_{j=1}^N 3\beta_j\sqrt{2\log(|V^j||V^{j-1}|)} 
\end{align*}
and the last term is upper bounded by
\begin{align*}
	6B\sqrt{n}\sum_{j=1}^N \beta_j\sqrt{\log(|V^j|)} \leq 12B\sqrt{n}\sum_{j=1}^N (\beta_j-\beta_{j+1})\sqrt{\log(|V^j|)}
	\leq 12B\sqrt{n}\int_{\beta_{N+1}}^{\beta_0}\sqrt{\log\cN_2(\delta \G,\z)}d\delta
\end{align*}
Given any $\rho\in(0,\gamma)$, we let $N=\max\{j: \beta_j>2\rho\}$. Then $\beta_{N}<4\rho$ and $\beta_{N+1}>\rho$, and thus
\begin{align*}
	\En \sup_{g\in\G}\left[ \sum_{t=1}^{n} \epsilon_t \bta_t(\epsilon) \Big(g(\z_t(\epsilon))-\v[\epsilon,g]^0_t(\epsilon)\Big)  \right] \leq B \inf_{\rho\in(0,\gamma)} \left\{ 4\rho n + 12\sqrt{n}\int_{\rho}^\gamma \sqrt{\log\cN_2(\delta, \G,\z)}d\delta \right\} 
\end{align*}
This concludes the proof.
\end{proof}

\begin{proof}[\textbf{Proof of Lemma~\ref{lem:reverse_dudley_with_alpha}}]
The proof closely follows that of Lemma~\ref{lem:reverse_dudley}, except for the way we use Lemma~\ref{lem:finite_lemma_with_etas} to take advantage of the subtracted quadratic term. We also employ an $\ell_\infty$ notion of sequential cover, rather than $\ell_2$. To this end, let $V'$ be a sequential $\gamma$-cover of $\G$ on $\z$ in the $\ell_\infty$ sense, i.e.
$$\forall \epsilon,~~ \forall g\in\G,~~ \exists \v\in V' \mbox{~~~s.t.~~~}  \max_{t\in[n]}|g(\z_t(\epsilon))-\v_t(\epsilon)| \leq \gamma$$
As before, let $V=V'\cup \{\boldsymbol{0}\}$ and denote by $\v[\epsilon,g]$ a $\gamma$-close tree promised by the definition. As in \eqref{eq:dudley_initial_split}, for any $c\in[0,1]$,
\begin{align*}
	&\En \sup_{g\in\G}\left[ \sum_{t=1}^{n} 4\epsilon_t \bta_t(\epsilon) g(\z_t(\epsilon)) - g(\z_t(\epsilon))^2 - \alpha\bta_t(\epsilon)^2 \right] \\
	&\leq \En \sup_{g\in\G}\left[ \sum_{t=1}^{n} \left\{4\epsilon_t \bta_t(\epsilon)\Big(g(\z_t(\epsilon))-\v[\epsilon,g]_t(\epsilon)\Big) - \frac{\alpha}{2}\bta_t(\epsilon)^2 \right\}- \sum_{t=1}^{n} \Big(g(\z_t(\epsilon))^2-c^2\v[\epsilon,g]_t(\epsilon)^2\Big) \right]  \\
	&~~~~~+ \En \max_{\v\in V'}\left[\sum_{t=1}^n 4\epsilon_t\bta_t(\epsilon)\v_t(\epsilon) - c^2\v_t(\epsilon)^2 -\frac{\alpha}{2}\bta_t(\epsilon)^2\right]
\end{align*}
Following the proof of Lemma~\ref{lem:reverse_dudley}, we claim that for any $\epsilon,g$ there exists an element $\v[\epsilon,g]\in V$ such that for any $t\in[n]$,
\begin{align}
	\label{eq:desired_l2_norm_relation_linfty}
	|g(\z_t(\epsilon)) | \geq c|\v[\epsilon,g]_t(\epsilon)|
\end{align}
First consider the easy case $\|g(\z_t(\epsilon))\|_\infty \leq C\gamma$, where $C=\frac{c}{1-c}$. Then $\boldsymbol{0}\in V$ provides a sequential $C\gamma$-cover in the $\ell_\infty$ sense. If, on the other hand, $\|g(\z_t(\epsilon))\|_\infty > C\gamma$, we fix any tree $\v[\epsilon,g]\in V$ that is $\gamma$-close in the $\ell_\infty$ sense to $g$ on the path $\epsilon$. With the same argument as in the proof of Lemma~\ref{lem:reverse_dudley}, we conclude \eqref{eq:desired_l2_norm_relation_linfty}.

Hence,
\begin{align}
	&\En \sup_{g\in\G}\left[ \sum_{t=1}^{n} 4\epsilon_t\bta_t(\epsilon) g(\z_t(\epsilon)) - g(\z_t(\epsilon))^2  - \alpha\bta_t(\epsilon)^2 \right] \notag \\
	&\leq 4\En \sup_{g\in\G}\left[ \sum_{t=1}^{n} \epsilon_t \bta_t(\epsilon)\Big(g(\z_t(\epsilon))-\v[\epsilon,g]_t(\epsilon)\Big) - \frac{\alpha}{2}\bta_t(\epsilon)^2 \right] \label{eq:dud_eq_split_no_square_a} \\ 
	& + \En \max_{\v\in V'}\left[\sum_{t=1}^{n}4\epsilon_t\bta_t(\epsilon)\v_t(\epsilon) - (1/4)\v_t(\epsilon)^2 -\frac{\alpha}{2}\bta_t(\epsilon)^2\right] \label{eq:term2a}
\end{align}
By Lemma~\ref{lem:finite_lemma_with_etas}, the term \eqref{eq:term2a} is upper bounded as
\begin{align}
	\label{eq:term_from_gamma_cover}
	\En_\epsilon \max_{\v\in V'}\left[\sum_{t=1}^{n}4\epsilon_t\bta_t(\epsilon)\v_t(\epsilon) - (1/4)\v_t(\epsilon)^2 - \frac{\alpha}{2}\bta_t(\epsilon)^2 \right] \leq \alpha^{-1} 16A^2 \log \cN_\infty(\gamma, \G,\z)
\end{align}
As for the term in \eqref{eq:dud_eq_split_no_square_a}, we write 
\begin{align*}
	&\En \sup_{g\in\G}\left[ \sum_{t=1}^{n} \epsilon_t \bta_t(\epsilon)\Big(g(\z_t(\epsilon))-\v[\epsilon,g]^0_t(\epsilon)\Big) - \frac{\alpha}{2}\bta_t(\epsilon)^2 \right] \\
	&= \En \sup_{g\in\G}\left[ \sum_{t=1}^{n} \epsilon_t \bta_t(\epsilon) \Big(g(\z_t(\epsilon))-\v[\epsilon,g]^N_t(\epsilon)\Big) - \frac{\alpha}{4}\bta_t(\epsilon)^2 \right.\\
	&\left.\hspace{2cm}+  \sum_{t=1}^{n} \sum_{j=1}^N \left\{ \epsilon_t \bta_t(\epsilon)\Big(\v[\epsilon,g]^j_t(\epsilon)-\v[\epsilon,g]^{j-1}_t(\epsilon)\Big) - \frac{\alpha}{4N}\bta_t(\epsilon)^2 \right\}\right] \\
	&\leq \En \sup_{g\in\G}\left[ \sum_{t=1}^{n} \epsilon_t \bta_t(\epsilon)\Big(g(\z_t(\epsilon))-\v[\epsilon,g]^N_t(\epsilon)\Big)  - \frac{\alpha}{4}\bta_t(\epsilon)^2 \right] \\
	&\hspace{2cm}+ \sum_{j=1}^N \En \sup_{g\in\G}\left[ \sum_{t=1}^{n} \epsilon_t\bta_t(\epsilon) \Big(\v[\epsilon,g]^j_t(\epsilon)-\v[\epsilon,g]^{j-1}_t(\epsilon)\Big) - \frac{\alpha}{4N}\bta_t(\epsilon)^2 \right]
\end{align*}
Using Cauchy-Schwartz inequality along with $ab\leq (1/2)(a^2+b^2)$,
{\small
\begin{align*}
	\frac{1}{n}\sum_{t=1}^{n} \epsilon_t \bta_t(\epsilon)\Big(g(\z_t(\epsilon))-\v[\epsilon,g]^N_t(\epsilon)\Big) 
	&\leq \sum_{t=1}^{n} \left( \frac{\sqrt{\alpha}\epsilon_t \bta_t(\epsilon)}{\sqrt{2n}} \right) \left(\sqrt{\frac{2}{n\alpha}}\Big(g(\z_t(\epsilon))-\v[\epsilon,g]^N_t(\epsilon)\Big)\right)  \\
	&\leq  \frac{1}{4n} \sum_{t=1}^{n} \alpha \bta_t(\epsilon)^2  +\sum_{t=1}^{n} \frac{1}{n\alpha}\Big(g(\z_t(\epsilon))-\v[\epsilon,g]^N_t(\epsilon)\Big)^2 \\
\end{align*}
}
and thus
$$\En \sup_{g\in\G}\left[ \sum_{t=1}^{n} \epsilon_t \bta_t(\epsilon)\Big(g(\z_t(\epsilon))-\v[\epsilon,g]^N_t(\epsilon)\Big)  - \frac{\alpha}{4}\bta_t(\epsilon)^2 \right] \leq \alpha^{-1}\beta_N n$$
where $\beta_j = 2^{-j}\gamma$. For the $j$-th link in the chain, recall that we can define
\begin{align*}
\w^{(s,r)}_t(\epsilon) = 
        \begin{cases} 
        \v^s_t(\epsilon)-\v^r_t(\epsilon) & \text{if there exists } g\in\G \mbox{ s.t. } \v^s = \v[g,\epsilon]^{j}, \v^r = \v[g, \epsilon]^{j-1} \\
        0 &\text{otherwise.}
        \end{cases}
\end{align*}
for all $t\in [n]$ and $\epsilon\in\{\pm1\}^n$. Then for any $j\in[N]$ and $\epsilon$,
\begin{align*}
	\sup_{g\in\G}\left[ \sum_{t=1}^{n} \epsilon_t \bta_t(\epsilon)\Big(\v[\epsilon,g]^j_t(\epsilon)-\v[\epsilon,g]^{j-1}_t(\epsilon)\Big) - \frac{\alpha}{4N}\bta_t(\epsilon)^2\right] \leq \max_{\w\in W}\left[ \sum_{t=1}^{n} \epsilon_t \bta_t(\epsilon)\w_t(\epsilon) - \frac{\alpha}{4N}\bta_t(\epsilon)^2\right]
\end{align*}
and it must hold by the definition of the cover that
$$|\w_t(\epsilon)| \leq 2\beta_j $$
for any $\w\in W^j$ and any path $\epsilon$ and any $t$. Putting everything together, and using Lemma~\ref{lem:finite_lemma_with_etas}, 
\begin{align*}
	&\En_\epsilon \sup_{g\in\G}\left[ \sum_{t=1}^{n} \epsilon_t \bta_t(\epsilon)\Big(g(\z_t(\epsilon))-\v[\epsilon,g]^0_t(\epsilon)\Big) - \frac{\alpha}{2}\bta_t(\epsilon)^2 \right] \leq \frac{n\beta_N}{\alpha} + \sum_{j=1}^N \frac{8N\beta_j^2}{\alpha}\log(|V^j||V^{j-1}|) 
\end{align*}
Simplifying and using $\beta_j = \beta_{j-1}-\beta_{j}$, we obtain an upper bound of
\begin{align*}
	\frac{n\beta_N}{\alpha} + \frac{16N}{\alpha}\sum_{j=1}^N \beta_j^2 \log(|V^j|) &= \frac{n\beta_N}{\alpha} + \frac{16N}{\alpha}\sum_{j=1}^N (\beta_{j-1}-\beta_j)\beta_j \log(|V^j|)\\
	&\leq \frac{n\beta_N}{\alpha} + \frac{16N}{\alpha}\int_{\beta_{N+1}}^{\beta_0}\delta \log\cN_\infty(\delta, \G,\z)d\delta
\end{align*}
Given any $\rho\in(0,\gamma)$, we let $N=\max\{j: \beta_j>2\rho\}$. Then $\beta_{N}<4\rho$ and $\beta_{N+1}>\rho$. Further, $N\leq \log(\gamma/\rho)$. Thus
\begin{align*}
	&\En \sup_{g\in\G}\left[ \sum_{t=1}^{n} \epsilon_t \bta_t(\epsilon) \Big(g(\z_t(\epsilon))-\v[\epsilon,g]^0_t(\epsilon)\Big) -\frac{\alpha}{2}\bta_t(\epsilon)^2 \right] \\ 
	&\leq \alpha^{-1}\inf_{\rho\in(0,\gamma)} \left\{ 4\rho n + 16\log(\gamma/\rho) \int_{\rho}^\gamma \delta \log\cN_\infty(\delta, \G,\z) d\delta \right\} 
\end{align*}
Together with \eqref{eq:term_from_gamma_cover} this concludes the proof.
\end{proof}

\begin{proof}[\textbf{Proof of Lemma~\ref{lem:lower_p_greater_2}}]
	Fix a $\beta>0$, and set $n=\fat_\beta(\F)$. Suppose $\x$ is an $\X$-valued tree of depth $n$ that is $\beta$-shattered by $\F$:
	$$\forall \epsilon, \exists f^\epsilon \in\F ~~~\mbox{s.t.}~~~ \epsilon_t (f^\epsilon(\x_t(\epsilon))-\bmu_t(\epsilon))\geq \beta/2$$
	where $\bmu$ is the witness to shattering. Since functions in $\F$ take values in $\left[-1,1\right]$, it is also the case that $\bmu$ is $\left[-1,1\right]$-valued, and thus $|f(\x_t(\epsilon))-\bmu_t(\epsilon)|\leq 2$ for all $f\in\F$. Then from \eqref{eq:lower_bd_rad_with_var} with the particular choices of $\x$ and $\bmu$ described above,
	\begin{align}
		V_n^0 &\geq \En \sup_{f\in\F}\left[ \sum_{t=1}^{n} 4\epsilon_t (f(\x_t(\epsilon))-\boldsymbol{\mu}_t(\epsilon)) - (f(\x_t(\epsilon))-\boldsymbol{\mu}_t(\epsilon))^2  \right] \\
		&\geq \En \sup_{f\in\F}\left[ \sum_{t=1}^{n} 4\epsilon_t (f(\x_t(\epsilon))-\boldsymbol{\mu}_t(\epsilon)) - 2|f(\x_t(\epsilon))-\boldsymbol{\mu}_t(\epsilon)|  \right] \\
		&\geq \En \left[ \sum_{t=1}^{n} 4\epsilon_t (f^\epsilon(\x_t(\epsilon))-\boldsymbol{\mu}_t(\epsilon)) - 2|f^\epsilon(\x_t(\epsilon))-\boldsymbol{\mu}_t(\epsilon)|  \right] 
	\end{align}
	Using the definition of shattering, we can further lower bound the above quantity by
	\begin{align*}
		&\En \left[ \sum_{t=1}^{n} 4|f^\epsilon(\x_t(\epsilon))-\boldsymbol{\mu}_t(\epsilon)| - 2|f^\epsilon(\x_t(\epsilon))-\boldsymbol{\mu}_t(\epsilon)|  \right] \geq \En \left[ \sum_{t=1}^{n} \beta  \right] 
		= n\beta
	\end{align*}
	Now, suppose $\fat_\beta(\F)=C/\beta^p$, $p>0$. Then $n=\fat_\beta(\F)$ implies $\beta = Cn^{-1/p}$. The result follows.
\end{proof}

\begin{proof}[\textbf{Proof of Lemma~\ref{lem:lower_p_less_2}}]
Assume $d=\fat_\beta(\F') \leq n$. Let $\z$ be an $\X$-valued tree of depth $d$ that is $\beta$-shattered by $\F'$ with a witness tree $\s$. Observe that the functions $f^\epsilon$ that guarantee 
\begin{align}
	\label{eq:fat_def}
	\forall t\in[n], ~\epsilon_t(f^\epsilon(\z_t(\epsilon))-\s_t(\epsilon))\geq \beta/2
\end{align}
do not necessarily take on values close to the $\s_t(\epsilon)\pm \beta/2$ interval. We augment $\F'$ with $2^d$ functions $g^\epsilon$ that take on the same values as $f^\epsilon$, except  \eqref{eq:fat_def} is satisfied with equality: $\epsilon_t(g^\epsilon(\z_t(\epsilon))-\s_t(\epsilon))= \beta/2$. Let $\F$ be the resulting class of functions, and $\G=\F\setminus\F'$. We now argue that $\fat_\beta(\F)$ cannot be more than $2d+4$, as we have only added at most $2^{d}$ functions to $\F'$. Suppose for the sake of contradiction that there exists a tree $\z$ of depth at least $2d+5$ shattered by $\F$. There must exist $2^{2d+5}$ functions that shatter $\z$ and only at most $2^{d}$ of them can be from $\G$. Let us label the leaves of $\z$ with the functions that shatter the corresponding path from the root; these functions are clearly distinct. Order the leaves of the tree in any way, and observe that there must exist a pair of functions from $\G$ with indices differing by at least $2^{d+4}$. It is easy to see that such two leaves can only have a common parent at $d+3$ levels from the leaves, and this yields a complete binary subtree of size $d+1$ that is shattered by functions in $\F'$, a contradiction.

We will now use the function class $\F$ to prove a lower bound. Recall that $\z$ is an $\X$-valued tree of depth $\fat_\beta$ that is $\beta$-shattered by $\G\subseteq\F$. Let $\s$ be the witness tree for the shattering. 
We will now show a construction of particular trees of depth 
\begin{align}
	\label{eq:larger_n}
	n' = \left\lceil \frac{n}{\fat_\beta} \right\rceil \fat_\beta
\end{align}
using the pair $\z,\s$.  Define $k = \lceil \frac{n}{\fat_\beta} \rceil = \frac{n'}{\fat_\beta}\geq 1$ and consider the $\X$-valued tree $\x$ and the $\reals$-valued tree $\boldsymbol{\mu}$ of depth $n'$ constructed as follows. For any path $\epsilon \in \{\pm 1\}^{n'}$ and any $t \in [n']$, set 
$$
\x_t(\epsilon) = \z_{\lceil \frac{t}{k}\rceil}\left(\tilde{\epsilon} \right), ~~~ \boldsymbol{\mu}_t(\epsilon) = \s_{\lceil \frac{t}{k}\rceil}\left(\tilde{\epsilon} \right)
$$
where $\tilde{\epsilon} \in \{\pm 1\}^{\fat_\beta}$  is the sequence of signs specified as 
$$
\tilde{\epsilon} = \left( \sign\left(\sum_{j=1}^k \epsilon_j\right) ,\sign\left(\sum_{j=k+1}^{2k} \epsilon_j\right), \ldots, \sign\left(\sum_{j= k \left(\fat_\beta - 1\right)}^{k\, \fat_\beta} \epsilon_j\right) \right) .
$$
We now lower bound \eqref{eq:lower_bd_rad_with_var} by choosing the particular $\x,\boldsymbol{\mu}$ defined above:
\begin{align*}
V^0_{n'} &\geq \En \sup_{f\in\F}\left[ \sum_{t=1}^{n'} 2\epsilon_t (f(\x_t(\epsilon))-\boldsymbol{\mu}_t(\epsilon)) - (f(\x_t(\epsilon))-\boldsymbol{\mu}_t(\epsilon))^2  \right]\\
&=\En \sup_{f\in\F}\left[ \sum_{t=1}^{n'} 2\epsilon_t (f(\z_{\lceil \frac{t}{k}  \rceil}(\tilde{\epsilon}))-\s_{\lceil \frac{t}{k}  \rceil}(\tilde{\epsilon})) - (f(\z_{\lceil \frac{t}{k}  \rceil}(\tilde{\epsilon}))-\s_{\lceil \frac{t}{k}  \rceil}(\tilde{\epsilon}))^2  \right] \ .
\end{align*}
Splitting the sum over $t$ into $\fat_\beta$ blocks, the above expression is equal to
\begin{align*}
&\En \sup_{f\in\F}\left[ \sum_{i=1}^{\fat_\beta}\sum_{j=(i-1)k + 1}^{i \cdot k} 2\epsilon_j (f(\z_i(\tilde{\epsilon}))-\s_i(\tilde{\epsilon})) - (f(\z_i(\tilde{\epsilon}))-\s_i(\tilde{\epsilon}))^2  \right] \\
&=\En \sup_{f\in\F}\left[ \sum_{i=1}^{\fat_\beta} 2(f(\z_i(\tilde{\epsilon}))-\s_i(\tilde{\epsilon})) \left(\sum_{j=(i-1)k + 1}^{i \cdot k} \epsilon_j \right) - k(f(\z_i(\tilde{\epsilon}))-\s_i(\tilde{\epsilon}))^2  \right] \\
&=\En \sup_{f\in\F}\left[ \sum_{i=1}^{\fat_\beta} 2\tilde{\epsilon}_i(f(\z_i(\tilde{\epsilon}))-\s_i(\tilde{\epsilon})) \left|\sum_{j=(i-1)k + 1}^{i \cdot k} \epsilon_j\right|  - k(f(\z_i(\tilde{\epsilon}))-\s_i(\tilde{\epsilon}))^2  \right]
\end{align*}
where the last step follows by the definition of $\tilde{\epsilon}$. Recall that $\z$ is shattered by the subset $\G$ and that the functions in $\G$ stay close to the witness tree $\s$. 
We obtain a lower bound
\begin{align*}
	\En \sup_{g\in\G}\left[ \sum_{i=1}^{\fat_\beta} 2\tilde{\epsilon}_i (f(\z_i(\tilde{\epsilon}))-\s_i(\tilde{\epsilon})) \left|\sum_{j=(i-1)k + 1}^{i \cdot k} \epsilon_j\right|  - k(f(\z_i(\tilde{\epsilon}))-\s_i(\tilde{\epsilon}))^2  \right] &\geq \En \sum_{i=1}^{\fat_\beta} \left( \beta \left|\sum_{j=(i-1)k + 1}^{i \cdot k} \epsilon_j\right|  - \frac{k\beta^2}{4}  \right) \\
	&\geq \fat_\beta(\F) \left( \beta \sqrt{\frac{k}{2}} - \frac{k\beta^2}{4} \right)
\end{align*}
where we used Khinchine's inequality in the last step. By the definition of $k$,
\begin{align*}
	\fat_\beta(\F)\beta \sqrt{\frac{k}{2}} 
 =  \fat_\beta(\F)\beta \sqrt{\frac{n'}{2\, \fat_\beta(\F)}} =  \frac{1}{\sqrt{2}} \beta \sqrt{n' \fat_\beta(\F)} 
\end{align*}
and
$$\fat_\beta(\F)\frac{k\beta^2}{4} = \frac{1}{4} n'\beta^2$$
We conclude that
\begin{align}
	\label{eq:lb1}
	V^0_{n'} \geq \frac{1}{4} \left(2\sqrt{2} \beta \sqrt{n' \fat_\beta(\F)} - n'\beta^2\right)
\end{align}

Now suppose $\fat_\beta(\F) = c/\beta^p$ for some $c>0$. First, we need to ensure that $\fat_{\beta}(\F) = c/\beta^p \leq n'$, as required by our construction. This means that $\beta \geq (cn')^{-1/p}$.
Plugging in the rate of $\fat_{\beta}(\F)$ into \eqref{eq:lb1},
$$2\sqrt{2} \beta \sqrt{n' \fat_\beta(\F)} - n'\beta^2 = 2\sqrt{2} c^{1/2}\beta^{1-p/2} \sqrt{n'}  - n'\beta^2$$
Observe that the setting of $\beta = (32c)^{1/(2+p)}(n')^{-1/(p+2)}$ yields a lower bound of
$$c_p \cdot (n')^{\frac{p}{p+2}}$$
where $c_p$ denotes a constant that may depend on $p$, and whose value may change from line to line. 

Examining \eqref{eq:unrolled_value_alpha_0}, we see that $V^0_{n}$ is nondecreasing with $n$. To illustrate this, let $n'>n$. For $t\in\{n+1,\ldots, n'\}$, we may choose $p_t$ in \eqref{eq:unrolled_value_alpha_0} as a delta distribution on $f^*(x_t)$, for any sequence of $x_t$, where $f^*$ is an optimal function over steps $\{1,\ldots, n\}$. Clearly, $V^0_{n'}\geq V^0_{n}$. In view of \eqref{eq:larger_n} and the above discussion, $V^0_{n'}\leq V^0_{2n-1}$, and thus
$$V^0_{2n}\geq V^0_{2n-1} \geq V^0_{n'} \geq c_p n^{\frac{p}{p+2}} \ .$$
\end{proof}

\begin{proof}[\textbf{Proof of Lemma \ref{lem:condradrel}}]
First note that when $t=n$ the initial condition is trivially satisfied as
$$
\Rad_n(x_{1:n},y_{1:n}) = \sup_{f\in\F}\left\{- \sum_{j=1}^n (f(x_j) - y_j)^2  \right\} = - \inf_{f\in\F}\sum_{j=1}^n (f(x_j) - y_j)^2  \ .
$$ 
Let us denote $$\widehat{L}_t(f) = \sum_{j=1}^t (f(x_j) - y_j)^2$$
and $$A_{t+1}(f) = \sum_{j=t+1}^{n} B\epsilon_j (f(\x_j(\epsilon))-\boldsymbol{\mu}_j(\epsilon)) - (f(\x_j(\epsilon))-\boldsymbol{\mu}_j(\epsilon))^2$$ 
To check admissibility note that we need to check the inequality in Equation \eqref{eq:admis}. To do so note that for any $x_t \in \X,  p_t \in \Delta([-B,B])$, 
{\small \begin{align*}
&\Es{y_t \sim p_t}{\left(\Es{y_t \sim p_t}{y_t} - y_t\right)^2} + \Es{y_t \sim p_t}{\Rad_{n}\left(x_{1:t},y_{1:t}\right)} = \Es{y_t \sim p_t}{  \left(\Es{y_t \sim p_t}{y_t} - y_t\right)^2 + \sup_{\x,\bmu}\En_\epsilon \sup_{f\in\F}\left\{ A_{t+1}(f) - \widehat{L}_t(f)  \right\} } 
\end{align*}}
Expanding the square in the first term and then the loss of $f$ at time $t$, we obtain 
{\small
\begin{align*}
&\En_{y_t \sim p_t}\Bigg[  \left(\Es{y_t \sim p_t}{y_t}\right)^2 - 2 y_t \Es{y_t \sim p_t}{y_t} + y_t^2 + \sup_{\x,\bmu}\En_\epsilon \sup_{f\in\F}\Big\{ A_{t+1}(f) - \widehat{L}_t(f)  \Big\} \Bigg] \\
& = \En_{y_t \sim p_t}\Bigg[  \left(\Es{y_t \sim p_t}{y_t}\right)^2 - 2 y_t \Es{y_t \sim p_t}{y_t}  + \sup_{\x,\bmu}\En_\epsilon \sup_{f\in\F}\Big\{ A_{t+1}(f)  - f^2(x_t) + 2 f(x_t) y_t - \widehat{L}_{t-1}(f)  \Big\} \Bigg] 
\end{align*}}
Rearranging, the above is equal to
{\small
\begin{align*}
&\En_{y_t \sim p_t}\Bigg[   \sup_{\x,\bmu}\En_\epsilon \sup_{f\in\F}\Big\{ A_{t+1}(f) - \widehat{L}_{t-1}(f) + 2 \left(\Es{y_t \sim p_t}{y_t}\right)^2 - f^2(x_t) - \left(\Es{y_t \sim p_t}{y_t}\right)^2 + 2 (f(x_t) - \Es{y_t \sim p_t}{y_t}) y_t  \Big\} \Bigg] \\
& = \En_{y_t \sim p_t}\Bigg[   \sup_{\x,\bmu}\En_\epsilon \sup_{f\in\F}\Big\{ A_{t+1}(f) - \widehat{L}_{t-1}(f) + 2 \left(\Es{y_t \sim p_t}{y_t}\right)^2  -\left(f(x_t) - \Es{y_t \sim p_t}{y_t}\right)^2 - 2 f(x_t) \Es{y_t \sim p_t}{y_t}   + 2 \left(f(x_t) - \Es{y_t \sim p_t}{y_t}\right) y_t  \Big\} \Bigg] 
\end{align*}}
which is 
{\small
\begin{align*}
&\En_{y_t \sim p_t}\Bigg[   \sup_{\x,\bmu}\En_\epsilon \sup_{f\in\F}\Big\{ A_{t+1}(f) - \widehat{L}_{t-1}(f)   - 2 \left( \Es{y_t \sim p_t}{y_t}\right)^2 \\
& ~~~~~~~~~~~~~~~~~~~~~~~~~~~~~~~~~~~~~~~~~~~~~~~~  -\left(f(x_t) - \Es{y_t \sim p_t}{y_t}\right)^2 - 2 \left(f(x_t) - \Es{y_t \sim p_t}{y_t} \right) \Es{y_t \sim p_t}{y_t}   + 2 \left(f(x_t) - \Es{y_t \sim p_t}{y_t}\right) y_t  \Big\} \Bigg] \\
& \le \En_{y_t \sim p_t}\Bigg[   \sup_{\x,\bmu}\En_\epsilon \sup_{f\in\F}\Big\{ A_{t+1}(f) - \widehat{L}_{t-1}(f)  -\left(f(x_t) - \Es{y_t \sim p_t}{y_t}\right)^2    + 2 \left(f(x_t) - \Es{y_t \sim p_t}{y_t}\right) \left(y_t -\Es{y_t \sim p_t}{y_t}\right) \Big\} \Bigg] 
\end{align*}
By Jensen's inequality, the above can be upper bounded by
\begin{align*}
&\En_{y_t, y'_t \sim p_t}\Bigg[   \sup_{\x,\bmu}\En_\epsilon \sup_{f\in\F}\Big\{ A_{t+1}(f) - \widehat{L}_{t-1}(f)  -\left(f(x_t) - \Es{y_t \sim p_t}{y_t}\right)^2    + 2 \left(f(x_t) - \Es{y_t \sim p_t}{y_t}\right) \left(y_t - y'_t\right) \Big\} \Bigg] \\
& = \En_{y_t, y'_t \sim p_t, \epsilon_t}\Bigg[   \sup_{\x,\bmu}\En_\epsilon \sup_{f\in\F}\Big\{ A_{t+1}(f) - \widehat{L}_{t-1}(f)  -\left(f(x_t) - \Es{y_t \sim p_t}{y_t}\right)^2    + 2 \epsilon_t \left(f(x_t) -  \Es{y_t \sim p_t}{y_t}\right) \left(y_t - y'_t\right) \Big\} \Bigg] 
\end{align*}
Since the inequalities above hold for any $x_t \in \X,  p_t \in \Delta([-B,B])$, we have
\begin{align*}
&\sup_{x_t \in \X,  p_t \in \Delta([-B,B])} \left[ \Es{y_t \sim p_t}{\left(\Es{y_t \sim p_t}{y_t} - y_t\right)^2} + \Es{y_t \sim p_t}{\Rad_{n}\left(x_{1:t},y_{1:t}\right)} \right]\\
&\leq \sup_{x_t \in \X,  p_t \in \Delta([-B,B])} \En_{y_t, y'_t \sim p_t, \epsilon_t}\Bigg[   \sup_{\x,\bmu}\En_\epsilon \sup_{f\in\F}\Big\{ A_{t+1}(f) - \widehat{L}_{t-1}(f)  -\left(f(x_t) - \Es{y_t \sim p_t}{y_t}\right)^2    + 2 \epsilon_t \left(f(x_t) -  \Es{y_t \sim p_t}{y_t}\right) \left(y_t - y'_t\right) \Big\} \Bigg] \\
&\le \sup_{\substack{x_t \in \X\\ y_t,y'_t ,\mu_t \in [-B,B]}}\En_{\epsilon_t}\Bigg[   \sup_{\x,\bmu}\En_\epsilon \sup_{f\in\F}\Big\{ A_{t+1}(f) - \widehat{L}_{t-1}(f)  -\left(f(x_t) - \mu_t\right)^2    + 2 \epsilon_t \left(f(x_t) -  \mu_t\right) (y_t- y'_t) \Big\} \Bigg] \\
& \le \sup_{\substack{x_t \in \X\\ y_t ,\mu_t \in [-B,B]}}\En_{\epsilon_t}\Bigg[   \sup_{\x,\bmu}\En_\epsilon \sup_{f\in\F}\Big\{ A_{t+1}(f) - \widehat{L}_{t-1}(f)   -\left(f(x_t) - \mu_t\right)^2    + 4 \epsilon_t \left(f(x_t) -  \mu_t\right) y_t \Big\} \Bigg] 
\end{align*}
Since the above is convex in $y_t$, we can replace the supremum over $[-B,B]$ to supremum over $\{-B,B\}$
\begin{align*}
&\sup_{\substack{x_t \in \X, \mu_t \in [-B,B]\\ y_t \in \{-B,B\}}}\En_{\epsilon_t}\Bigg[   \sup_{\x,\bmu}\En_\epsilon \sup_{f\in\F}\Big\{ A_{t+1}(f) - \widehat{L}_{t-1}(f)    -\left(f(x_t) - \mu_t\right)^2    + 4 \epsilon_t \left(f(x_t) -  \mu_t\right) y_t \Big\} \Bigg] \\
& = \sup_{\substack{x_t \in \X\\ \mu_t \in [-B,B]}}\En_{\epsilon_t}\Bigg[   \sup_{\x,\bmu}\En_\epsilon \sup_{f\in\F}\Big\{ A_{t+1}(f) - \widehat{L}_{t-1}(f)    -\left(f(x_t) - \mu_t\right)^2    + 4 B \epsilon_t \left(f(x_t) -  \mu_t\right)  \Big\} \Bigg] \\
& = \sup_{\x,\bmu}\En_\epsilon\Bigg[    \sup_{f\in\F}\Big\{ \sum_{j=t}^{n} B\epsilon_j (f(\x_j(\epsilon))-\boldsymbol{\mu}_j(\epsilon)) - (f(\x_j(\epsilon))-\boldsymbol{\mu}_j(\epsilon))^2 - \widehat{L}_{t-1}(f)  \Bigg]  = \Rad_{n}\left(x_{1:t-1},y_{1:t-1}\right)
\end{align*}}
Thus we have shown that $\Rad_n$ is an admissible relaxation. Further,  $(\hat{y} - y_t)^2 + \Rad_{n}\left(x_{1:t},(y_{1:t-1},y_t)\right)$ is a convex function of $y_t$ and so for the estimator one can use 
$$
\hat{y}_t = \frac{\Rad_{n}(x_{1:t},(y_{1:t-1},B)) - \Rad_{n}(x_{1:t},(y_{1:t-1},-B)) }{4B}
$$ 
(no clipping is needed above as $\hat{y}_t$ is always between $-B$ and $B$). For the above estimator one enjoys the regret bound
$$
\Reg_n \le  \Rad_n(\cdot)
$$
Note that this is exactly the bound in Eq. \eqref{eq:val_upper_0_alpha}.
\end{proof}

\begin{proof}[\textbf{Proof of Proposition \ref{prop:recipe}}]
Notice that the above recipe closely follows the notion of relaxation provided in \citep{rakhlin2012relax}. All we need to do is check that the relaxation derived satisfies admissibility and initial conditions. By Step 1 of the recipe, since the offset Rademacher relaxation is admissible to start with, the derived relaxation also satisfies initial condition. To show admissibility condition notice that the set $[-B,B]$ is compact and convex and $(\hat{y}_t - y_t)^2 + \Rel{n}{x_{1:t},y_{1:t}}$ is a convex function of $\hat{y}_t$. Hence applying minimax theorem, we see that,
\begin{align*}
&\inf_{\hat{y}_t \in [-B,B]}  \sup_{y_t \in [-B,B]}\left\{ (\hat{y}_t - y_t)^2 + \Rel{n}{x_{1:t},y_{1:t}} \right\} \\
& = \sup_{p_t \in \Delta([-B,B])} \inf_{\hat{y}_t}\left\{ \Es{y_t \sim p_t}{(\hat{y}_t - y_t)^2 + \Rel{n}{x_{1:t},y_{1:t}}} \right\}\\
&  = \sup_{p_t \in \Delta([-B,B])}\left\{ \inf_{\hat{y}_t}\ \Es{y_t \sim p_t}{(\hat{y}_t - y_t)^2} + \Es{y_t \sim p_t}{\Rel{n}{x_{1:t},y_{1:t}}} \right\}\\
&  = \sup_{p_t \in \Delta([-B,B])}\left\{  \Es{y_t \sim p_t}{\left(\Es{y_t \sim p_t}{y_t} - y_t\right)^2} + \Es{y_t \sim p_t}{\Rel{n}{x_{1:t},y_{1:t}}} \right\}
\end{align*}
Hence the admissibility condition can be rewritten as :
\begin{align}\label{eq:admis}
\forall x_t \in \X,~~~~ \sup_{p_t \in \Delta([-B,B])}\left\{  \Es{y_t \sim p_t}{\left(\Es{y_t \sim p_t}{y_t} - y_t\right)^2} + \Es{y_t \sim p_t}{\Rel{n}{x_{1:t},y_{1:t}}} \right\} \le \Rel{n}{x_{1:t-1},y_{1:t-1}} 
\end{align}
\end{proof}

\begin{proof}[\textbf{Proof of Corollary \ref{cor:exp}}]
As done in \citep{rakhlin2012relax} for the case of finite class of experts, in the Rademacher relaxation one can replace the $\max_{f \in \F}$ with a limit of soft-max as follows:
{\small \begin{align*}
\Rad_n(x_{1:t},y_{1:t}) &=  \sup_{\x,\bmu}\En_\epsilon \max_{f\in\F}\left[ \sum_{j=t+1}^{n} 4 B\epsilon_j (f(\x_j(\epsilon))-\boldsymbol{\mu}_j(\epsilon)) - (f(\x_j(\epsilon))-\boldsymbol{\mu}_j(\epsilon))^2 - \sum_{j=1}^t (f(x_j) - y_j)^2  \right] \\
& =  \sup_{\x,\bmu}\En_\epsilon \inf_{\lambda > 0} \lambda^{-1} \log\left(\sum_{f\in\F}\exp\left( \lambda \sum_{j=t+1}^{n} 4 B\epsilon_j (f(\x_j(\epsilon))-\boldsymbol{\mu}_j(\epsilon)) - (f(\x_j(\epsilon))-\boldsymbol{\mu}_j(\epsilon))^2 - \lambda \sum_{j=1}^t (f(x_j) - y_j)^2\right)  \right)\\
& \le  \inf_{\lambda > 0}\Bigg\{ \lambda^{-1} \log\left(\sum_{f\in\F} \exp\left(  - \lambda \sum_{j=1}^t (f(x_j) - y_j)^2\right) \right) \\
& ~~~~~~~~~~~~~~~+ \sup_{\x,\bmu}  \lambda^{-1} \log\left( \En_\epsilon\exp\left( \lambda \sum_{j=t+1}^{n} 4 B\epsilon_j (f(\x_j(\epsilon))-\boldsymbol{\mu}_j(\epsilon)) - (f(\x_j(\epsilon))-\boldsymbol{\mu}_j(\epsilon))^2\right) \right) \Bigg\}
\end{align*}}
Not notice that if we set $\lambda = B^{-2}$, the proof of Lemma~\ref{lem:finite_lemma_with_etas} exactly shows that
$$
 \sup_{\x,\bmu}  \lambda^{-1} \log\left( \En_\epsilon\exp\left( \lambda \sum_{j=t+1}^{n} 4 B\epsilon_j (f(\x_j(\epsilon))-\boldsymbol{\mu}_j(\epsilon)) - (f(\x_j(\epsilon))-\boldsymbol{\mu}_j(\epsilon))^2\right) \right)  \le B^2 \log |\F|
$$
Hence we arrive at our relaxation 
$$
\Rel{n}{x_{1:t},y_{1:t}} = B^2 \log\left(\sum_{f\in\F} \exp\left(  - B^{-2} \sum_{j=1}^t (f(x_j) - y_j)^2\right) \right) 
$$
Now to show admissibility, note that  
{\small \begin{align*}
\sup_{x_t, p_t}& \Es{y_t \sim p_t}{ (y_t - \E{y_t})^2 + \Rel{n}{x_{1:t},y_{1:t}}  } \\
& = \sup_{x_t, p_t} \Es{y_t \sim p_t}{ y^2_t  - (\E{y_t})^2 + B^2 \log\left(\sum_{f\in\F} \exp\left(  - B^{-2} \sum_{j=1}^t (f(x_j) - y_j)^2\right) \right)  } \\
& = \sup_{x_t, p_t}\Es{y_t \sim p_t}{ B^2\log\left(\exp\left(B^{-2}y^2_t - B^{-2} (\E{y_t})^2\right)\right) + B^2 \log\left(\sum_{f\in\F} \exp\left(  - B^{-2} \sum_{j=1}^t (f(x_j) - y_j)^2\right) \right)  } \\
& = \sup_{x_t, p_t}\Es{y_t \sim p_t}{  B^2 \log\left(\sum_{f\in\F} \exp\left( B^{-2} y^2_t - B^{-2} (\E{y_t})^2  - B^{-2} \sum_{j=1}^t (f(x_j) - y_j)^2\right) \right)  } \\
& = \sup_{x_t, p_t}\Es{y_t \sim p_t}{  B^2 \log\left(\sum_{f\in\F} \exp\left( - B^{-2}(\E{y_t})^2  + 2  B^{-2} f(x_t) y_t - B^{-2} f^2(x_t) - B^{-2} \sum_{j=1}^{t-1} (f(x_j) - y_j)^2\right) \right)  } \\
& = \sup_{x_t, p_t}\Es{y_t \sim p_t}{  B^2 \log\left(\sum_{f\in\F} \exp\left( - B^{-2}(\E{y_t} - f(x_t))^2  + 2  B^{-2} f(x_t) (y_t - \E{y_t})  - B^{-2} \sum_{j=1}^{t-1} (f(x_j) - y_j)^2\right) \right)  } \\
\end{align*}}
Now by convexity (see \cite{audibert2009fast}) we can take the expectation w.r.t. $y_t$ inside and hence we get,
\begin{align*}
\sup_{x_t, p_t}& \Es{y_t \sim p_t}{ (y_t - \E{y_t})^2 + \Rel{n}{x_{1:t},y_{1:t}}  } \\
& \le \sup_{x_t, p_t}\left\{ B^2 \log\left(\sum_{f\in\F} \exp\left( - B^{-2}(\E{y_t} - f(x_t))^2    - B^{-2} \sum_{j=1}^{t-1} (f(x_j) - y_j)^2\right) \right)  \right\}\\
& \le  B^2 \log\left(\sum_{f\in\F} \exp\left(  - B^{-2} \sum_{j=1}^{t-1} (f(x_j) - y_j)^2\right) \right) \\
& = \Rel{n}{x_{1:t-1},y_{1:t-1}}
\end{align*}
Again as we used above (see \cite{audibert2009fast}) we have that the relaxation is such that $(\hat{y} - y_t)^2 + \Rel{n}{x_{1:t},(y_{1:t-1},y_t)}$ is a convex function of $y_t$ and so the estimator is given by
\begin{align*}
\hat{y}_t & = \mrm{Clip}\left(\frac{\Rel{n}{x_{1:t},(y_{1:t-1},B)} - \Rel{n}{x_{1:t},(y_{1:t-1},-B)} }{4B}\right)\\
& = \mrm{Clip}\left(\frac{B}{4} \log\left(\frac{\sum_{f\in\F} \exp\left(  - B^{-2} \sum_{j=1}^{t-1} (f(x_j) - y_j)^2 - B^{-2} (f(x_t) - B)^2\right)}{\sum_{f\in\F} \exp\left(  - B^{-2} \sum_{j=1}^{t-1} (f(x_j) - y_j)^2  - B^{-2} (f(x_t) + B)^2\right)} \right)  \right)
\end{align*}

Now the final regret bound we obtain is given by $\Reg_n \le  \Rel{n}{\cdot}$ and so we conclude that
$$
\Reg_n \le B^2 \log\left| \F \right|
$$
\end{proof}

\begin{proof}[\textbf{Proof of Corollary \ref{cor:linreg}}]
For simplicity, each input instance $x_t \in \reals^d$ we define vector in $\reals^{d+1}$ as $z_t = (0,x_t)$, the vector obtained by concatenating $0$ before $x_t$. Further given trees $\x$ and $\mbf{\mu}$, we write the  $\z$ as the $[-B,B] \times \X$ valued tree corresponding to $\x$ and $\mbf{\mu}$ obtained by concatenating $\mu$'s before $x$'s on every node. Also for every linear predictor $f \in \F$ define corresponding $w = (-1,f)$. The unnormalized regret over the rounds $-d$ to $n$ can be written as
$$
\sum_{t=1}^n (\hat{y}_t - y_t)^2 - \inf_{w}\left\{ \sum_{t=1}^n (\ip{w}{z_t} - y_t)^2 + \lambda \norm{w}_2^2 \right\}
$$
Hence, we have,

{\small
\begin{align*}
& \Rad_n(x_{1:t},y_{1:t}) = \sup_{\z}\En_\epsilon \sup_{f\in\F}\left[ \sum_{j=t+1}^{n-1} 4 B\epsilon_j \ip{w}{\z_j(\epsilon)} - (\ip{w}{\z_j(\epsilon)})^2 - \sum_{j=1}^t (\ip{w}{z_j} - y_j)^2 - \lambda \norm{w}_2^2 \right]\\
& = 2  \sup_{\z}\En_\epsilon \sup_{w}\left[ \inner{w,\sum_{j=t+1}^{n} 2 B\epsilon_j  \z_j(\epsilon) + \sum_{j=1}^t y_j z_j} - \frac{1}{2}w^\top\left( \sum_{j=t+1}^{n} \z_j(\epsilon)  \z_j(\epsilon)^\top + \sum_{j=1}^t z_j z_j^\top + \lambda I\right)w^\top  \right] - \sum_{j=1}^t y_j^2 
\end{align*} }
Let us denote ${\bf A}_{t+1:n}(\z) = \sum_{j=t+1}^{n} \z_j(\epsilon)  \z_j(\epsilon)^\top$ and ${\bf B}_t = \sum_{j=1}^t z_j z_j^\top$.
Using Fenchel-Young inequality for 
$$\frac{1}{2}w^\top\left( {\bf A}_{t+1:n}(\z) + {\bf B}_t + \lambda I\right)w^\top$$ and its conjugate we get,
{\small
\begin{align*}
\Rad_n(x_{1:t},y_{1:t}) &\le    \sup_{\z}\En_\epsilon \norm{\sum_{j=t+1}^{n} 2 B\epsilon_j  \z_j(\epsilon) + \sum_{j=1}^t y_j z_j}^2_{\left({\bf A}_{t+1:n}(\z) + {\bf B}_t + \lambda I\right)^{-1}}  - \sum_{j=1}^t y_j^2
\end{align*}
The idea now is to obtain a further upper bound by removing the dependence on the tree $\z$. Opening the square with only the $n$-th term, the above expression is equal to
\begin{align*}
&\sup_{\z}\En_\epsilon\Bigg[ \norm{\sum_{j=t+1}^{n-1} 2 B\epsilon_j  \z_j(\epsilon) + \sum_{j=1}^t y_j z_j}^2_{\left({\bf A}_{t+1:n}(\z) + {\bf B}_t + \lambda I\right)^{-1}}  - \sum_{j=1}^t y_j^2 + 4 B^2 \z_n(\epsilon)^\top \left({\bf A}_{t+1:n}(\z) + {\bf B}_t + \lambda I\right)^{-1} \z_n(\epsilon)\Bigg] \\
\end{align*}
By the standard argument we may upper bound the quadratic terms by a ratio of determinants $\Delta$: 
$$\z_n(\epsilon)^\top \left({\bf A}_{t+1:n}(\z) + {\bf B}_t + \lambda I\right)^{-1} \z_n(\epsilon) \leq \left(1 - \frac{\Delta\left({\bf A}_{t+1:n-1}(\z) + {\bf B}_t + \lambda I\right)}{\Delta\left({\bf A}_{t+1:n}(\z) + {\bf B}_t + \lambda I\right)}\right)$$
Using the inequality $1-x\leq -\log (x)$ for $x>0$, we obtain an upper bound
\begin{align*}
&\sup_{\z}\Bigg\{\Es{\epsilon}{\norm{\sum_{j=t+1}^{n-1} 2 B\epsilon_j  \z_j(\epsilon) + \sum_{j=1}^t y_j z_j}^2_{\left({\bf A}_{t+1:n-1}(\z) + {\bf B}_t + \lambda I\right)^{-1}}}  - \sum_{j=1}^t y_j^2+ 4 B^2\Es{\epsilon}{\log\left(\frac{\Delta\left({\bf A}_{t+1:n}(\z) + {\bf B}_t + \lambda I\right)}{\Delta\left({\bf A}_{t+1:n-1}(\z) + {\bf B}_t + \lambda I\right)}\right)}\Bigg\}
\end{align*}
Proceeding in similar fashion by peeling off terms from the norm, we arrive at,
\begin{align*}
\Rad_n(x_{1:t},y_{1:t}) &\le  \norm{\sum_{j=1}^t y_j z_j}^2_{\left({\bf B}_t + \lambda I\right)^{-1}}  - \sum_{j=1}^t y_j^2+ 4 B^2\sup_{\z}\Es{\epsilon}{\log\left(\frac{\Delta\left({\bf A}_{t+1:n}(\z) + {\bf B}_t + \lambda I\right)}{\Delta\left({\bf B}_t + \lambda I\right)}\right)} \\
&\le \norm{\sum_{j=1}^t y_j z_j}^2_{\left({\bf B}_t + \lambda I\right)^{-1}} + 4 B^2 \log\left(\frac{\left(n/d\right)^d}{\Delta\left({\bf B}_t + \lambda I\right)}\right)  - \sum_{j=1}^t y_j^2
\end{align*}}
and we take this last expression as our relaxation $\Rel{n}{x_{1:t},y_{1:t}}$.
Now notice that since $z_t$'s are $0$ on the first coordinate, the relaxation can be rewritten as
\begin{align*}
\Rel{n}{x_{1:t},y_{1:t}} 
&= \norm{\sum_{j=1}^t y_j x_j}^2_{\left({\bf \tilde{B}}_t + \lambda I\right)^{-1}} - \sum_{j=1}^t y_j^2 + 4 B^2 \log\left(\frac{\left(n/d\right)^d}{\Delta\left({\bf B}_t + \lambda I\right)}\right) 
\end{align*}
where ${\bf \tilde{B}}_t = \sum_{j=1}^t x_j x_j^\top$. By conjugacy, the relaxation is equal to
\begin{align*}
&\sup_{f \in \F}\left\{2 \sum_{j=1}^t y_j \ip{f}{x_j} - f^\top\left({\bf \tilde{B}}_t + \lambda I\right) f \right\} - \sum_{j=1}^t y_j^2 + 4 B^2 \log\left(\frac{\left(n/d\right)^d}{\Delta\left({\bf B}_t + \lambda I\right)}\right)  \\
& = - \inf_{f \in \F}\left\{ \sum_{j=1}^t (f(x_j) - y_j)^2 + \lambda \norm{f}_2^2 \right\}+ 4 B^2 \log\left(\frac{\left(n/d\right)^d}{\Delta\left({\bf B}_t + \lambda I\right)}\right)  
\end{align*}
We now prove admissibility of relaxation as follows: 
{\small
\begin{align*}
\sup_{p_t} & \Es{y_t \sim p_t}{ (y_t - \E{y_t})^2 + \Rel{n}{x_{1:t},y_{1:t}} }\\
& = \sup_{p_t}   \Es{y_t \sim p_t}{ (y_t - \E{y_t})^2  - \inf_{f \in \F}\left\{\sum_{j=1}^t \left(\ip{f}{x_j} - y_j\right)^2 + \lambda \norm{f}^2 \right\}} + 4 B^2 \log\left(\frac{\left(n/d\right)^d}{\Delta\left({\bf B}_t + \lambda I\right)}\right) 
\end{align*}}
The first term, in view of \eqref{eq:diff_of_squares}, is equal to
{\small
\begin{align*}
&\sup_{p_t}   \Es{y_t \sim p_t}{     \sup_{f \in \F}\left\{- \sum_{j=1}^{t-1} \left(\ip{f}{x_j} - y_j\right)^2  - (\ip{f}{x_t} - \E{y_t})^2 + 2 (y_t - \E{y_t}) \left(\ip{f}{x_t} - \E{y_t}\right)  + \lambda \norm{f}^2 \right\}}  \\
& \le \sup_{\mu_t}   \Es{\epsilon_t}{     \sup_{f \in \F}\left\{- \sum_{j=1}^{t-1} \left(\ip{f}{x_j} - y_j\right)^2  - (\ip{f}{x_t} - \mu_t)^2 + 4 B \epsilon_t  \left(\ip{f}{x_t} - \mu_t\right)  + \lambda \norm{f}^2 \right\}} 
\end{align*}
and the inequality arises from symmetrization exactly as in the proof of of Lemma~\ref{lem:expansion}. Once again, rewriting the above using conjugacy and converting to the $z_t$ notation by appending a coordinate, the relaxation is upper bounded by 
\begin{align*}
&\sup_{z_t} \Es{\epsilon_t}{ \norm{\sum_{j=1}^{t-1}y_j z_j + 2 B \epsilon_t z_t}_{\left({\bf B}_{t-1} + z_t z_t^\top + \lambda I\right)^{-1}}^2} + 4 B^2 \log\left(\frac{\left(n/d\right)^d}{\Delta\left({\bf B}_t + \lambda I\right)}\right)  - \sum_{j=1}^{t-1} y_j^2\\
& =  \sup_{z_t}  \norm{\sum_{j=1}^{t-1}y_j z_j}_{\left({\bf B}_{t-1} + z_t z_t^\top + \lambda I\right)^{-1}}^2 + 4 B^2 z_t^\top \left({\bf B}_{t-1} + z_t z_t^\top + \lambda I\right)^{-1} z_t + 4 B^2 \log\left(\frac{\left(n/d\right)^d}{\Delta\left({\bf B}_t + \lambda I\right)}\right)  - \sum_{j=1}^{t-1} y_j^2
\end{align*}
which is further upper bounded by
\begin{align*}
&\sup_{z_t}  \norm{\sum_{j=1}^{t-1}y_j z_j}_{\left({\bf B}_{t-1}  + \lambda I\right)^{-1}}^2 + 4 B^2 \log\left(\frac{\Delta \left({\bf B}_{t} + \lambda I\right)}{\Delta \left({\bf B}_{t-1} + \lambda I\right)}\right) + 4 B^2 \log\left(\frac{\left(n/d\right)^d}{\Delta\left({\bf B}_t + \lambda I\right)}\right)  - \sum_{j=1}^{t-1} y_j^2\\
& = \norm{\sum_{j=1}^{t-1}y_j z_j}_{\left({\bf B}_{t-1}  + \lambda I\right)^{-1}}^2 + 4 B^2 \log\left(\frac{\left(n/d\right)^d}{\Delta\left({\bf B}_{t-1} + \lambda I\right)}\right)  - \sum_{j=1}^{t-1} y_j^2\\
& = \Rel{n}{x_{1:t-1},y_{1:t-1}}
\end{align*}}
Thus we have shown admissibility and further this relaxation is such that $(\hat{y} - y_t)^2 + \Rel{n}{x_{1:t},(y_{1:t-1},y_t)}$ is a convex function of $y_t$ and so the forecast associated with this relaxation is simply 
\begin{align*}
\hat{y}_t & = \mrm{Clip}\left(\frac{\norm{\sum_{j=1}^{t-1} y_j x_j + B x_t}^2_{\left({\bf B}_t + \lambda I\right)^{-1}}   - \norm{\sum_{j=1}^{t-1} y_j x_j -  B x_t}^2_{\left({\bf B}_t + \lambda I\right)^{-1}}   }{4B}\right)
\end{align*}
Expanding out the two norm square terms we conclude that
\begin{align*}
\hat{y}_t & = \mrm{Clip}\left( x_t^\top \left({\bf B}_t + \lambda I\right)^{-1}  \left(\sum_{j=1}^{t-1} y_j x_j\right)
\right)
\end{align*}
Notice that this is exactly the clipped version of the Vovk-Azoury-Warmuth forecaster. The final regret bound we obtain is given by $\Reg \le  \Rel{n}{\cdot}$ and so we conclude that for any $f \in \F$, regret against this linear predictor is bounded as :
\begin{align*}
\frac{1}{n}\sum_{t=1}^n (\hat{y}_t - y_t)^2 \le \frac{1}{n}  \sum_{t=1}^n (f^\top x_t - y_t)^2 + \frac{\lambda}{2 n} \norm{f}_2^2 + \frac{4 d B^2 \log\left(\frac{n}{\lambda d}\right)}{n}
\end{align*}

\end{proof}

\end{document}